\documentclass{article}

\usepackage{arxiv}

\usepackage[utf8]{inputenc} 
\usepackage[T1]{fontenc}    
\usepackage{hyperref}       
\usepackage{amsmath}
\usepackage{amsthm}
\usepackage{amsfonts}       
\usepackage{microtype}      
\usepackage{listings}
\usepackage{graphicx}
\usepackage{wrapfig}
\usepackage{booktabs}
\usepackage{multirow}
\usepackage{pdflscape}
\usepackage[table]{xcolor}
\usepackage{flafter}
\usepackage{placeins}
\usepackage{tikz}
\usetikzlibrary{arrows.meta,calc,intersections}


\newtheorem{theorem}{Theorem}
\newtheorem{proposition}[theorem]{Proposition}

\theoremstyle{remark}

\title{Geometry-Aware Diffusion Guidance via Curvature-Adaptive Tubular Correction}
\author{
 Enze Jiang \\
  School of Mathematical Sciences\\
  Shanghai Jiao Tong University\\
  Shanghai, China \\
  \texttt{geguojia@sjtu.edu.cn} \\
   \And
 Jinwei He \\
  School of Mathematical Sciences\\
  Shanghai Jiao Tong University\\
  Shanghai, China \\
  \texttt{hjw10086@sjtu.edu.cn} \\
   \And
 Zheng Ma \\
  School of Mathematical Sciences\\
  Shanghai Jiao Tong University\\
  Shanghai, China \\
  \texttt{zhengma@sjtu.edu.cn} \\
}

\begin{document}
\maketitle
\begin{abstract}
Gradient-guided diffusion samplers provide flexible priors for inverse problems and conditional generation, but strong guidance can move the sampling trajectory into regions where the learned score is poorly supported. Existing tangent-projection strategies limit first-order departure from an iso-density surface, yet discard potentially useful normal motion and overlook the second-order departure induced by tangent motion on a curved surface. We introduce curvature-adaptive tubular correction (CAT), a training-free plugin that regulates both effects within a shared, noise-dependent geometric budget. CAT decomposes the guidance gradient into normal and tangent components, charges normal displacement at first order and tangent displacement according to directional curvature, and obtains their jointly optimal magnitudes from a one-dimensional dual equation. Armijo backtracking calibrates the resulting finite step against the actual guidance objective, while matrix-free directional derivatives avoid constructing the full score Jacobian. We establish local guarantees for the tubular approximation, uniqueness of the correction, and sufficient objective decrease. Across seven inverse problems on FFHQ and ImageNet, CAT improves the evaluated pixel- and latent-space host samplers, with particularly consistent gains in perceptual metrics. It also improves black hole reconstruction on InverseBench and yields the lowest FID among the compared methods at every tested classifier-free guidance scale, while maintaining stable saturation and contrast. These results support curvature-aware tubular control as a reusable mechanism for stabilizing diffusion guidance.
\end{abstract}

\section{Introduction}
\label{sec:introduction}

\begin{wrapfigure}[20]{r}{0.50\textwidth}
  \centering
  \begingroup
  \newcommand{\LabelFontSize}{14.2}
  \newcommand{\RegionFontSize}{16.2}
  \newcommand{\VectorLineWidth}{1.25pt}
  \newcommand{\ManifoldLineWidth}{1.32pt}
  \newcommand{\BoundaryLineWidth}{0.70pt}
  \newcommand{\NormalLineWidth}{1.05pt}
  \newcommand{\ArrowLength}{3.7mm}
  \newcommand{\ArrowWidth}{3.0mm}
  \newcommand{\PointRadius}{0.16cm}
  
  \newcommand{\UpperNormalLen}{120}
  \newcommand{\LowerNormalLen}{500}
  \newcommand{\RightTangentProbeLen}{1400}
  \newcommand{\LowerNormalProbeLen}{1400}
  
  \definecolor{TubeFill}{HTML}{F6F2FC}
  \definecolor{TubeBoundary}{HTML}{B6A0D7}
  \definecolor{ManifoldPurple}{HTML}{5F4B8B}
  \definecolor{UpdatePurple}{HTML}{7033A4}
  \definecolor{PriorBlue}{HTML}{0D5CAF}
  \definecolor{DPSRed}{HTML}{C52D26}
  \definecolor{GuideCircle}{HTML}{8F9AAF}
  \resizebox{\linewidth}{!}{%
    \begin{tikzpicture}[
      x=0.01cm,y=-0.01cm,
      line cap=round,line join=round,
      every node/.style={inner sep=0pt,outer sep=0pt},
      label/.style={
        anchor=base west,
        font=\fontsize{\LabelFontSize}{17}\selectfont
      },
      region/.style={
        label,
        font=\fontsize{\RegionFontSize}{19}\selectfont
      },
      vector/.style={
        line width=\VectorLineWidth,
        -{Stealth[length=\ArrowLength,width=\ArrowWidth]}
      },
      boundary/.style={
        draw=TubeBoundary,
        line width=\BoundaryLineWidth,
        dash pattern=on 4.5pt off 4.5pt
      }
    ]
    
    \coordinate (Xt)     at (156,300);
    \coordinate (Anchor) at (643,442);
    \coordinate (Next)   at (1163,495);
    \coordinate (RawTip) at (1000,750);
    
    \coordinate (TopLeft)  at (17,413);
    \coordinate (TopC1)    at (312,285);
    \coordinate (TopC2)    at (608,204);
    \coordinate (TopMid)   at (879,251);
    \coordinate (TopC3)    at (1147,289);
    \coordinate (TopC4)    at (1458,424);
    \coordinate (TopRight) at (1807,554);
    
    \coordinate (BottomLeft)  at (146,782);
    \coordinate (BottomC1)    at (394,640);
    \coordinate (BottomC2)    at (684,564);
    \coordinate (BottomMid)   at (912,600);
    \coordinate (BottomC3)    at (1160,620);
    \coordinate (BottomC4)    at (1415,777);
    \coordinate (BottomRight) at (1622,879);
    
    \def\TubeInset{40}
    
    \foreach \pname in {TopLeft,TopC1,TopC2,TopMid,TopC3,TopC4,TopRight}{
      \coordinate (\pname) at ($(\pname)+(0,\TubeInset)$);
    }
    \foreach \pname in {BottomLeft,BottomC1,BottomC2,BottomMid,BottomC3,BottomC4,BottomRight}{
      \coordinate (\pname) at ($(\pname)+(0,-\TubeInset)$);
    }
    
    \coordinate (ManifoldLeft)  at (83,596);
    \coordinate (ManifoldC1)    at (288,515);
    \coordinate (ManifoldC2)    at (476,458);
    \coordinate (UpdateC1)      at ($(Anchor)+(194,-15)$);
    \coordinate (UpdateC2)      at ($(Next)+(-204,-57)$);
    \coordinate (ManifoldC5)    at ($(Next)+(215,64)$);
    \coordinate (ManifoldC6)    at (1570,662);
    \coordinate (ManifoldRight) at (1707,709);
    
    \pgfmathsetlengthmacro{\UpperNormalDist}{\UpperNormalLen*0.01cm}
    \pgfmathsetlengthmacro{\LowerNormalDist}{\LowerNormalLen*0.01cm}
    \pgfmathsetlengthmacro{\RightTangentProbeDist}{\RightTangentProbeLen*0.01cm}
    \pgfmathsetlengthmacro{\LowerNormalProbeDist}{\LowerNormalProbeLen*0.01cm}
    
    \coordinate (UpdateNormalTop) at
      ($(Anchor)!\UpperNormalDist!90:(UpdateC1)$);
    \coordinate (UpdateNormalBottom) at
      ($(Anchor)!\LowerNormalDist!-90:(UpdateC1)$);
    
    \coordinate (TangentLeft)  at ($(Anchor)!-1!(UpdateC1)$);
    \coordinate (TangentRight) at ($(Anchor)!2.4!(UpdateC1)$);
    
    \coordinate (UpdateTangentProbe) at
      ($(Anchor)!\RightTangentProbeDist!(UpdateC1)$);
    \coordinate (UpdateNormalProbe) at
      ($(Anchor)!\LowerNormalProbeDist!-90:(UpdateC1)$);
    
    \coordinate (LabelXt)           at ($(Xt)+(-36,-35)$);
    \coordinate (LabelAnchor)       at ($(Anchor)+(-25,-40)$);
    \coordinate (LabelNaivePoint)   at ($(RawTip)+(18,18)$);
    \coordinate (LabelManifold)     at (1540,600);
    
    \path[use as bounding box] (0,0) rectangle (1808,1026);
    \clip (0,0) rectangle (1808,1026);
    \fill[white] (0,0) rectangle (1808,1026);
    
    \path[fill=TubeFill]
      (TopLeft) .. controls (TopC1) and (TopC2) .. (TopMid)
                .. controls (TopC3) and (TopC4) .. (TopRight)
      -- (BottomRight)
                .. controls (BottomC4) and (BottomC3) .. (BottomMid)
                .. controls (BottomC2) and (BottomC1) .. (BottomLeft)
      -- cycle;
    
    \draw[boundary,name path=TopBoundary]
      (TopLeft) .. controls (TopC1) and (TopC2) .. (TopMid)
                .. controls (TopC3) and (TopC4) .. (TopRight);
    
    \draw[boundary,name path=BottomBoundary]
      (BottomLeft) .. controls (BottomC1) and (BottomC2) .. (BottomMid)
                   .. controls (BottomC3) and (BottomC4) .. (BottomRight);
    
    \draw[ManifoldPurple,line width=\ManifoldLineWidth]
      (ManifoldLeft) .. controls (ManifoldC1) and (ManifoldC2) .. (Anchor)
      .. controls (UpdateC1) and (UpdateC2) .. (Next)
      .. controls (ManifoldC5) and (ManifoldC6) .. (ManifoldRight);
    
    \draw[GuideCircle,draw opacity=0.38,line width=1.05pt,dash pattern=on 5pt off 4pt]
      (TangentLeft) -- (TangentRight);
    
    \path[name path=RightTangentProbePath]
      (Anchor) -- (UpdateTangentProbe);
    \path[name path=LowerNormalProbePath]
      (Anchor) -- (UpdateNormalProbe);
    
    \path[name intersections={of=RightTangentProbePath and TopBoundary, by=TangentHit}];
    \path[name intersections={of=LowerNormalProbePath and BottomBoundary, by=NormalHit}];

    
    
    \path let
      \p1 = (RawTip),
      \p2 = (BottomMid),
      \p3 = (BottomC3),
      \p4 = (BottomC4),
      \p5 = (BottomRight)
    in
      \pgfextra{
        \xdef\CircleCX{\the\dimexpr\x1\relax}
        \xdef\CircleCY{\the\dimexpr\y1\relax}
    
        \xdef\BXzero{\the\dimexpr\x2\relax}
        \xdef\BYzero{\the\dimexpr\y2\relax}
        \xdef\BXone{\the\dimexpr\x3\relax}
        \xdef\BYone{\the\dimexpr\y3\relax}
        \xdef\BXtwo{\the\dimexpr\x4\relax}
        \xdef\BYtwo{\the\dimexpr\y4\relax}
        \xdef\BXthree{\the\dimexpr\x5\relax}
        \xdef\BYthree{\the\dimexpr\y5\relax}
      };
    
    \pgfmathdeclarefunction{boundaryX}{1}{%
      \pgfmathparse{%
        (1-#1)^3*(\BXzero/1pt)
        +3*(1-#1)^2*#1*(\BXone/1pt)
        +3*(1-#1)*#1^2*(\BXtwo/1pt)
        +#1^3*(\BXthree/1pt)
      }%
    }
    
    \pgfmathdeclarefunction{boundaryY}{1}{%
      \pgfmathparse{%
        (1-#1)^3*(\BYzero/1pt)
        +3*(1-#1)^2*#1*(\BYone/1pt)
        +3*(1-#1)*#1^2*(\BYtwo/1pt)
        +#1^3*(\BYthree/1pt)
      }%
    }
    
    \pgfmathdeclarefunction{boundaryDistSq}{1}{%
      \pgfmathparse{%
        (boundaryX(#1)-\CircleCX/1pt)^2
        +(boundaryY(#1)-\CircleCY/1pt)^2
      }%
    }
    
    \pgfmathsetmacro{\SearchA}{0}
    \pgfmathsetmacro{\SearchB}{1}
    \pgfmathsetmacro{\GoldenRatio}{0.61803398875}
    
    \foreach \i in {1,...,45}{%
      \pgfmathsetmacro{\SearchC}
        {\SearchB-\GoldenRatio*(\SearchB-\SearchA)}
      \pgfmathsetmacro{\SearchD}
        {\SearchA+\GoldenRatio*(\SearchB-\SearchA)}
    
      \pgfmathsetmacro{\DistC}{boundaryDistSq(\SearchC)}
      \pgfmathsetmacro{\DistD}{boundaryDistSq(\SearchD)}
    
      \ifdim\DistC pt<\DistD pt
        \global\let\SearchB\SearchD
      \else
        \global\let\SearchA\SearchC
      \fi
    }
    
    \pgfmathsetmacro{\TangentParam}{(\SearchA+\SearchB)/2}
    
    \pgfmathsetmacro{\TangentX}{boundaryX(\TangentParam)}
    \pgfmathsetmacro{\TangentY}{boundaryY(\TangentParam)}
    
    \coordinate (ProjectedHit) at
      (\TangentX pt,\TangentY pt);
    \coordinate (ProjectedFoot) at ($(Anchor)!(ProjectedHit)!(UpdateC1)$);
      
    
    
    \draw[UpdatePurple,line width=0.95pt,dash pattern=on 4pt off 3pt]
      (ProjectedFoot) -- (ProjectedHit);
    
    \draw[vector,UpdatePurple,line width=1.05pt,dash pattern=on 4pt off 3pt]
      (Anchor) -- (ProjectedFoot);
    
    \draw[vector,UpdatePurple,line width=1.05pt,dash pattern=on 4pt off 3pt]
      (ProjectedFoot) -- (ProjectedHit);
    
    \path let
      \p1 = (RawTip),
      \p2 = (ProjectedHit)
    in
      \pgfextra{
        \pgfmathsetlengthmacro{\TempRadius}
          {veclen(\x2-\x1,\y2-\y1)}
        \global\let\NaiveCircleRadius\TempRadius
      };
    
    \draw[
      GuideCircle,
      draw opacity=0.38,
      line width=0.95pt,
      dash pattern=on 5pt off 4pt
    ]
      (RawTip) circle[radius=\NaiveCircleRadius];
    
    \draw[vector,UpdatePurple,line width=1.35pt]
      (Anchor) -- (ProjectedHit);
    
    \filldraw[fill=UpdatePurple,draw=white,line width=0.85pt]
      (ProjectedHit) circle[radius=\PointRadius];
    
    \node[label,text=UpdatePurple,anchor=north]
      at ($(ProjectedHit)+(0,25)$)
      {$x_t^{-}\ (\mathrm{CAT})$};
    
    \draw[vector,PriorBlue,shorten >=6pt]
      (Xt) -- (Anchor);
    \draw[vector,DPSRed]
      (Anchor) -- (RawTip);
    
    \filldraw[fill=red,draw=white,line width=0.85pt]
      (RawTip) circle[radius=\PointRadius];
    
    \filldraw[fill=PriorBlue,draw=white,line width=0.85pt]
      (Xt) circle[radius=\PointRadius];
    \filldraw[fill=black,draw=white,line width=0.95pt]
      (Anchor) circle[radius=\PointRadius];
    
    \node[label] at (LabelXt) {$x_{t+\Delta t}^{-}$};
    \node[label] at (LabelAnchor) {$x_t^{+}$};
    
    \node[label,text=DPSRed] at (LabelNaivePoint) {$x_t^{-}\ (\mathrm{host})$};
    
    \node[region,text=ManifoldPurple] at (LabelManifold)
      {$\mathcal{M}_t$};
    
    \begin{scope}[shift={(1160,55)}]
      \draw[vector,DPSRed] (0,0) -- (105,0);
      \node[label] at (130,8) {Host update};

      \draw[vector,UpdatePurple] (0,55) -- (105,55);
      \node[label] at (130,63) {Host+CAT update};

      \draw[vector,PriorBlue] (0,110) -- (105,110);
      \node[label] at (130,118) {Unconditional generation};

      \path[fill=TubeFill,draw=TubeBoundary,line width=\BoundaryLineWidth]
        (0,145) rectangle (105,180);
      \node[label] at (130,178) {Safe region};
    \end{scope}
    
    \draw[black,line width=0.28pt]
      (0.5,0.5) rectangle (1807.5,1025.5);
    
    \end{tikzpicture}
  }%
  \endgroup
  \caption{CAT within one reverse-diffusion step. The host update can leave the safe region, whereas CAT uses curvature-adaptive normal--tangent allocation to produce $x_t^{-}$ inside it.}
  \label{fig:cat_overview}
\end{wrapfigure}
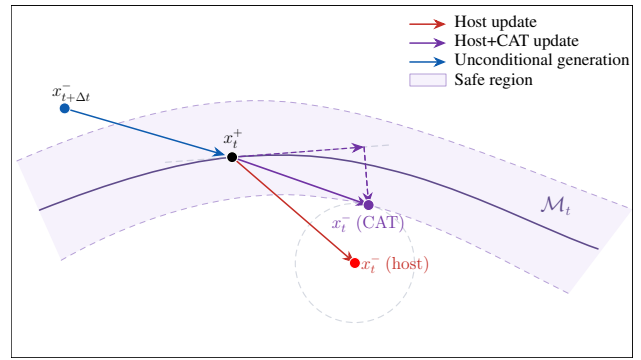

Diffusion models are powerful priors for recovering images from incomplete, blurred, or nonlinear measurements because a single pretrained model can be reused across forward operators without task-specific retraining \cite{ho2020denoising,song2021scorebased,kawar2022denoising,wang2023zeroshot}. Their impact increasingly extends to scientific and medical inverse problems, including PDE-constrained inversion\cite{jiang2025odedps} and full-waveform inversion \cite{peng2026robust,min2026decoupled}, accelerated MRI \cite{chung2022scoremri}, and 3D CT/MRI reconstruction \cite{chung2023diffusionmbir}. Gradient-guided posterior samplers exploit this flexibility by inserting a differentiable measurement objective into reverse diffusion \cite{chung2023diffusion}. Their central difficulty is to enforce the observation strongly enough to resolve the inverse problem without driving the evolving sample away from regions well represented by the learned prior.

\clearpage

This difficulty is geometric as well as numerical. Guided states far from the high-probability noisy-data region receive comparatively little supervision during score training, so aggressive measurement updates can expose poorly constrained score estimates and accumulate artifacts. Manifold-constrained and on-manifold methods reduce this risk \cite{chung2022improving,he2024manifold}, and score-based projection obtains a locally tangent update by removing the score-parallel component of the measurement gradient \cite{hamidi2025score}. Exact tangent projection, however, can remove measurement information carried by the normal component. Moreover, tangency preserves an iso-density surface only to first order: on a curved surface, a finite tangent step creates a second-order normal departure. Neither a single isotropic step size nor a hard tangent projection accounts for both effects.

We address this problem with \emph{curvature-adaptive tubular correction} (CAT), a training-free plugin for gradient-guided diffusion solvers. CAT replaces exact surface preservation with a noise-dependent tube around the local iso-density surface. At each reverse step, it decomposes the guidance gradient into normal and tangent components and allocates a shared departure budget between them: normal displacement consumes tube thickness at first order, whereas tangent displacement is charged at second order according to its directional curvature. This formulation retains useful normal guidance while suppressing geometrically expensive motion in either component.

The resulting correction is both tractable and explicitly controlled. We formulate the component allocation as a convex subproblem with a unique solution obtained from a scalar dual equation. Because the tube is a local approximation and the initial host step may still be too large, we apply Armijo backtracking to require sufficient decrease of the actual guidance objective while preserving feasibility of the second-order tube model. In implementation, CAT needs only a directional curvature computed by automatic differentiation, rather than the full score Jacobian, and all operations are performed in the host solver's working variable, enabling the same construction in pixel and latent space. Figure~\ref{fig:cat_overview} summarizes the resulting one-step geometry.

We evaluate CAT on seven inverse problems over FFHQ and ImageNet by attaching it to the pixel-space DPS sampler and the latent-space PSLD and ReSample samplers. CAT improves each evaluated host--task pair, with especially consistent reductions in LPIPS and FID, and remains effective across sweeps of guidance strength, tubular radius, and measurement noise. Component ablations show that Armijo backtracking alone, first-order tubular control, and tangent-only projection each fall short of the full method. Beyond natural-image restoration, DPS+CAT improves image fidelity over its host on InverseBench black hole imaging at both observation times; in classifier-free text-to-image generation, CFG+CAT attains the lowest FID at every tested guidance scale while keeping saturation and contrast stable. Our contributions are:
\begin{itemize}
  \item We introduce a second-order tubular view of diffusion guidance that jointly controls first-order normal departure and curvature-induced tangent departure without discarding either component of the guidance gradient.
  \item We derive a unique scalar-dual correction, pair it with Armijo finite-step calibration, and establish local guarantees for tubular approximation and measurement descent.
  \item We provide a matrix-free, training-free implementation and validate its reuse across pixel- and latent-space inverse solvers, scientific imaging, and classifier-free guidance through broad comparisons, robustness analyses, and targeted ablations.
\end{itemize}

\section{Preliminaries}
\label{sec:preliminaries}

\subsection{Diffusion Models}
\label{sec:diffusion_models}

Diffusion models represent the data prior through a Gaussian perturbation process $x_t=\mu_tx_0+\sigma_t\epsilon$, where $\epsilon\sim\mathcal N(0,I_n)$, and learn a score model $s_\theta(x,t)\approx\nabla_x\log p_t(x)$ to reverse this process \cite{ho2020denoising,song2021scorebased}. For the continuous-time forward SDE $dx=f(x,t)dt+g(t)dw_t$, the reverse drift is $f(x,t)-g(t)^2\nabla_x\log p_t(x)$; replacing the exact score with $s_\theta$ yields a practical generative sampler. The same score provides the Tweedie estimate
\begin{equation}
  \widehat x_{0,\theta}(x_t,t)
  =
  \frac{
    x_t+\sigma_t^2s_\theta(x_t,t)
  }{\mu_t}
  \approx
  \mathbb E[x_0\mid x_t]
  \label{eq:tweedie_estimate}
\end{equation}
which maps the noisy diffusion state to the clean domain where the measurement operator acts.

\subsection{Diffusion Posterior Sampling}
\label{sec:dps}

Diffusion posterior sampling (DPS) recovers $x_0$ from measurements $y=A(x_0)+\xi$, where $A:\mathbb R^n\rightarrow\mathbb R^m$ is a known differentiable operator and $\xi\sim\mathcal N(0,\sigma_y^2I_m)$. It augments the pretrained prior score with measurement guidance through $\nabla_{x_t}\log p_t(x_t\mid y)=\nabla_{x_t}\log p_t(x_t)+\nabla_{x_t}\log p_t(y\mid x_t)$ \cite{chung2023diffusion}. Because the conditional likelihood at time $t$ is generally intractable, DPS evaluates it at the Tweedie estimate, $p_t(y\mid x_t)\approx p(y\mid\widehat x_{0,\theta}(x_t,t))$. Under Gaussian measurement noise, this yields the differentiable guidance objective
\begin{equation}
  L_t(x)
  =
  \frac{1}{2\sigma_y^2}
  \left\|
    A\!\left(\widehat x_{0,\theta}(x,t)\right)-y
  \right\|_2^2.
  \label{eq:dps_loss}
\end{equation}

At each reverse time, DPS first produces a prior anchor and then applies a measurement-gradient correction:
\begin{equation}
  x_t^+=\Psi^{\mathrm{prior}}_{t\leftarrow t+\Delta t}(x_{t+\Delta t}^-),
  \qquad
  q_t=\nabla_xL_t(x_t^+),
  \qquad
  x_t^-=x_t^+-\alpha_tq_t,
  \label{eq:dps_anchor}
\end{equation}
where $\alpha_t>0$ controls the guidance strength. Standard DPS scales every component of $q_t$ uniformly, without distinguishing first-order normal departure from curvature-induced tangent departure. CAT introduces this geometric distinction while retaining the host sampler's reverse transition.

\section{Related Work}
\label{sec:related_work}

\paragraph{Training-free diffusion priors for inverse problems.} A broad class of methods repurposes an unconditional diffusion model as a task-independent image prior and enforces measurements only at inference time.  For linear operators, DDRM exploits a spectral decomposition of the forward model, whereas DDNM separates its range and null spaces to impose data consistency during reverse diffusion \cite{kawar2022denoising,wang2023zeroshot}.  DPS instead differentiates a likelihood surrogate evaluated at the Tweedie estimate, which accommodates noisy nonlinear operators without task-specific training \cite{chung2023diffusion}; DiffPIR places a diffusion denoiser inside a plug-and-play restoration scheme \cite{zhu2023denoising}.  More recent approaches improve the surrounding sampling mechanism: DPnP alternates a likelihood-only proximal sampler with a diffusion-prior sampler and provides robustness guarantees \cite{xu2024provably}, while DAPS decouples successive noise levels to permit larger transitions and correct errors accumulated earlier in the trajectory \cite{zhang2025improving}.  These works primarily determine how the likelihood and diffusion prior form a conditional sampler.  Our work addresses a complementary question: once a differentiable measurement gradient is available, how should its normal and tangent components be scaled so that the correction remains in a locally reliable region of the learned prior?  Consequently, our correction can be attached to gradient-guided solvers without replacing their underlying reverse transition.

\paragraph{Geometry-preserving diffusion guidance.} The closest line of work controls conditional updates using the geometry of the diffusion prior.  MCG identifies accumulated off-manifold error as a failure mode of diffusion inverse solvers and introduces a manifold-constrained gradient correction \cite{chung2022improving}.  MPGD develops training-free on-manifold guidance using pretrained autoencoders and shows that a corresponding shortcut preserves the latent manifold for latent diffusion models \cite{he2024manifold}.  Score-based manifold projection directly treats the score as the normal to an iso-density surface and removes the score-parallel component of the measurement gradient, thereby producing a first-order tangent update \cite{hamidi2025score}.  These methods motivate preserving manifold fidelity, but enforcing an exactly on-manifold or tangent correction can discard useful measurement motion in the normal direction and does not account for the second-order departure of a tangent step on a curved surface.  We instead admit both components inside a signed-distance tube: normal motion consumes the budget at first order, tangent motion consumes it through directional curvature, and a scalar dual equation selects their unique jointly optimal scaling.  This tubular formulation therefore relaxes exact projection while still preventing unrestricted movement into regions where the learned score is poorly supported.

\paragraph{Inverse problems with latent diffusion models.} Latent diffusion priors reduce generative cost but make measurement consistency more delicate because the forward operator is composed with a nonlinear decoder.  ReSample addresses this difficulty by solving a hard data-consistency subproblem and resampling the resulting estimate back to the noisy latent manifold \cite{song2024solving}.  P2L jointly adapts the text prompt, latent code, and pixel reconstruction through alternating optimization \cite{chung2024prompttuning}, whereas STSL constructs a tractable surrogate for second-order Tweedie information to improve latent posterior sampling \cite{rout2024beyond}.  These methods redesign data consistency or the posterior approximation specifically for latent diffusion.  In contrast, our construction is representation-agnostic: the decoder is included in the measurement objective, while the score geometry, gradient decomposition, curvature, and tubular step are all computed with respect to the current working variable.  The same correction can thus operate in pixel or latent space without changing its geometric derivation.

\section{Our Method}
\label{sec:method}

\subsection{Motivation}
\label{sec:method_motivation}

Strong measurement guidance can move diffusion trajectories beyond regions well represented during score training, leading to accumulated prior error and visible texture or color artifacts. The concentration result of Chung~et~al.~\cite{chung2022improving} characterizes the noisy-data region underlying this behavior.

\begin{proposition}[Concentration of noisy data \cite{chung2022improving}]
\label{prop:noisy_concentration}
Suppose the clean distribution is uniform on a locally linear $n$-dimensional manifold $\mathcal M\subset\mathbb R^D$, where $n<D$. Under $x_t=\mu_t x_0+\sigma_t\epsilon$, the noisy marginal concentrates around the $(D-1)$-dimensional shell
\begin{equation}
  \mathcal S_t=\left\{x\in\mathbb R^D:
  \operatorname{dist}(x,\mu_t\mathcal M)=\sigma_t\sqrt{D-n}\right\}.
\end{equation}
\end{proposition}

Because score training draws noisy states from $p_t$, supervision is concentrated near $\mathcal S_t$, while aggressive guidance can carry an iterate into less reliable regions and amplify score error over subsequent reverse steps. CAT directly addresses this failure mode by regulating each correction within a noise-dependent tube around the local prior geometry. Unlike exact manifold projection, this tubular design preserves useful measurement motion while controlling both normal displacement and curvature-induced tangent departure.

\subsection{Curvature-adaptive tubular (CAT) correction}
\label{sec:method_correction}

CAT applies to any diffusion solver that provides a differentiable guidance gradient $q_t=\nabla L_t(x_t)$ in its working space, whether $x_t$ is a noisy image or a latent code. We present CAT on top of DPS, with $x_t$ denoting the prior anchor $x_t^+$ in \eqref{eq:dps_anchor} and $\alpha_t>0$ the host guidance step size. The same construction transfers directly to other gradient-guided samplers.

CAT uses the iso-density surface through the current state to describe local prior geometry. Let $\phi_t=\log p_t$, $c_t:=\phi_t(x_t)=\log p_t(x_t)$, and $\mathcal M_t:=\{x\in\mathbb R^D:\phi_t(x)=c_t\}$. The surface $\mathcal M_t$ is updated at every reverse step. When $\phi_t$ is $C^3$ with nonzero gradient near $x_t$, its local component is a regular hypersurface oriented by the score-induced unit normal $\nu_t(z):=\nabla\phi_t(z)/\|\nabla\phi_t(z)\|_2$.
Let $\mathcal U_t$ be a tubular neighborhood with a unique closest-point projection onto $\mathcal M_t$. Its signed distance, oriented by $\nu_t$, is
\begin{equation}
  r(x,\mathcal M_t)
  :=
  \begin{cases}
    +\operatorname{dist}(x,\mathcal M_t),
    & x\text{ lies on the side pointed to by }\nu_t,\\
    -\operatorname{dist}(x,\mathcal M_t),
    & x\text{ lies on the opposite side},
  \end{cases}
  \label{eq:signed_distance_definition}
\end{equation}
where $\operatorname{dist}(x,\mathcal M_t)=\inf_{z\in\mathcal M_t}\|x-z\|_2$, so $|r(x,\mathcal M_t)|=\operatorname{dist}(x,\mathcal M_t)$. For $z\in\mathcal M_t$, we have $r(z,\mathcal M_t)=0$ and $\nabla_xr(z,\mathcal M_t)=\nu_t(z)$. Assuming $r$ is $C^3$ on $\mathcal U_t$, the local score geometry at $x_t$ is
\begin{equation}
  s_t=\nabla\phi_t(x_t),\quad
  g_t=\|s_t\|_2,\quad
  H_t=\nabla^2\phi_t(x_t),\quad
  \nu_t:=\nu_t(x_t)=\frac{s_t}{g_t},\quad
  P_{N,t}=\nu_t\nu_t^\top,\quad
  P_{T,t}=I-P_{N,t},
  \label{eq:local_geometry}
\end{equation}
where $P_{N,t}$ and $P_{T,t}$ project onto the score-normal and tangent spaces, respectively, and $T_{x_t}\mathcal M_t=\{v:s_t^\top v=0\}$. CAT permits guidance within the noise-adaptive tube $\mathcal T_t(R_t):=\{x\in\mathcal U_t:|r(x,\mathcal M_t)|\leq R_t\}$ with $R_t=\rho\sigma_t$, where $\rho>0$ controls the geometric budget. Scaling the radius with $\sigma_t$ follows the width of the noisy distribution and progressively tightens guidance during denoising.

\begin{theorem}[Second-order tubular expansion]
\label{thm:tubular_expansion}
Assume the preceding local regularity conditions and $\sup_x\|D^3r(x,\mathcal M_t)\|_{\mathrm{op}}\leq M_3$ in the relevant tubular neighborhood. The local curvature matrix is
\begin{equation}
  B_t
  =
  \frac{1}{g_t}P_{T,t}H_tP_{T,t}.
  \label{eq:curvature_matrix}
\end{equation}
For every sufficiently local displacement $d=d_N+d_T$, where $d_N=P_{N,t}d$ and $d_T=P_{T,t}d$,
\begin{equation}
  r(x_t+d,\mathcal M_t)
  =
  \nu_t^\top d_N
  +\frac{1}{2}d_T^\top B_td_T
  +R_3(d),
  \qquad
  |R_3(d)|
  \leq
  \frac{M_3}{6}\|d\|_2^3.
  \label{eq:tubular_expansion}
\end{equation}
\end{theorem}

Theorem~\ref{thm:tubular_expansion} reveals the key asymmetry exploited by CAT: normal motion consumes tube thickness at first order, whereas tangent motion is charged at second order according to directional curvature. For component magnitudes $r_N$ and $r_T$ along a unit tangent direction $u$, let $K_t=|u^\top B_tu|$. The resulting non-canceling budget $r_N+\frac{1}{2}K_tr_T^2\leq R_t$ conservatively controls the second-order signed-distance model without allowing the two components to cancel. Appendix~\ref{app:geometry_proof} provides the derivation.

We next allocate the shared budget optimally between the normal and tangent guidance components. Decompose $q_{N,t}=P_{N,t}q_t$ and $q_{T,t}=P_{T,t}q_t$, with magnitudes $a_t=\|q_{N,t}\|_2$ and $b_t=\|q_{T,t}\|_2$. For each nonzero component, define $e_{N,t}=q_{N,t}/a_t$ and $e_{T,t}=q_{T,t}/b_t$; vanishing components receive zero displacement. We seek a descent correction $d_t=-r_{N,t}e_{N,t}-r_{T,t}e_{T,t}$ and measure tangent cost by the directional curvature $K_t=|e_{T,t}^\top B_te_{T,t}|$, with $K_t=0$ when $b_t=0$. The optimal magnitudes solve
\begin{equation}
\begin{aligned}
  \min_{r_{N,t},r_{T,t}\geq0}\quad&
  -a_tr_{N,t}-b_tr_{T,t}
  +\frac{r_{N,t}^2+r_{T,t}^2}{2\alpha_t}\\
  \text{subject to}\quad&
  r_{N,t}+\frac{1}{2}K_tr_{T,t}^2\leq R_t.
\end{aligned}
\label{eq:trust_region_problem}
\end{equation}
The objective stays close to the host update while maximizing first-order measurement descent. Without the tube constraint, its solution $(r_{N,t},r_{T,t})=(\alpha_ta_t,\alpha_tb_t)$ exactly recovers the original isotropic guidance step.

\begin{theorem}[Unique curvature-adaptive step]
\label{thm:kkt_solution}
For $\alpha_t,R_t>0$ and $a_t,b_t,K_t\geq0$, \eqref{eq:trust_region_problem} has a unique global minimizer given by $r_{N,t}(\lambda_t)=\alpha_t(a_t-\lambda_t)_+$ and $r_{T,t}(\lambda_t)=\alpha_tb_t/(1+\alpha_t\lambda_tK_t)$, where $\lambda_t=0$ if $\alpha_ta_t+\frac{1}{2}K_t(\alpha_tb_t)^2\leq R_t$. Otherwise, $\lambda_t$ is the unique positive root of
\begin{equation}
  F_t(\lambda_t)
  :=
  \alpha_t(a_t-\lambda_t)_+
  +\frac{K_t}{2}
  \left(
    \frac{\alpha_tb_t}
    {1+\alpha_t\lambda_tK_t}
  \right)^2
  =R_t.
  \label{eq:dual_equation}
\end{equation}
\end{theorem}

The scalar equation is inexpensive to solve independently for each sample using bracketing and bisection. The resulting component-wise step sizes are
\begin{equation}
  \eta_{N,t}
  =
  \alpha_t
  \left(1-\frac{\lambda_t}{a_t}\right)_+,
  \qquad
  \eta_{T,t}
  =
  \frac{\alpha_t}{1+\alpha_t\lambda_tK_t},
  \label{eq:adaptive_step_sizes}
\end{equation}
 giving $d_t=-\eta_{N,t}q_{N,t}-\eta_{T,t}q_{T,t}$. Thus, CAT adapts the normal component to its first-order tube cost and the tangent component to its directional curvature through a single scalar multiplier. Appendix~\ref{app:kkt_proof} provides the proof.

\begin{theorem}[Measurement descent and Armijo acceptance]
\label{thm:descent_armijo}
Suppose $\nabla L_t$ is $L_g$-Lipschitz along the update segment, and let $d_t^*$ solve \eqref{eq:trust_region_problem}. If $0<\alpha_t\leq1/L_g$, then $L_t(x_t+d_t^*)\leq L_t(x_t)$, with strict inequality when $q_t\neq0$. For any $\alpha_t>0$ and $q_t\neq0$, $d_t^*$ remains a strict descent direction, and Armijo backtracking with $c,\beta\in(0,1)$ terminates at a scale $\gamma_t\in\{1,\beta,\beta^2,\ldots\}$ satisfying
\begin{equation}
  L_t(x_t+\gamma_td_t^*)
  \leq
  L_t(x_t)
  +c\gamma_tq_t^\top d_t^*
  <
  L_t(x_t).
  \label{eq:armijo_condition}
\end{equation}
Moreover, the accepted displacement remains second-order feasible: $\gamma_tr_{N,t}^*+\frac{1}{2}K_t\gamma_t^2(r_{T,t}^*)^2\leq R_t$.
\end{theorem}

Theorem~\ref{thm:descent_armijo} couples geometric control with reliable finite-step optimization. Rather than estimating the trajectory-dependent constant $L_g$, CAT applies Armijo backtracking to each curvature-adaptive proposal. The accepted update achieves sufficient decrease in the actual guidance objective while preserving second-order tubular feasibility. To reduce the cost of repeated objective evaluations, we additionally introduce an Armijo period $p$: backtracking is performed once every $p$ reverse steps, and the most recently accepted scale $\gamma_t$ is reused at the intervening steps. Increasing $p$ therefore amortizes the line-search overhead and provides a controllable efficiency--quality trade-off, which we evaluate in Section~\ref{sec:efficiency_results}. Appendix~\ref{app:descent_proof} gives the proof of the per-step acceptance result.

Combining curvature-adaptive allocation with finite-step calibration, CAT replaces the host's isotropic correction by
\begin{equation}
  x_t^-
  =x_t+\gamma_td_t
  =x_t-\gamma_t
  \left(
    \eta_{N,t}q_{N,t}+\eta_{T,t}q_{T,t}
  \right).
  \label{eq:cat_one_step_update}
\end{equation}
Here the component-wise step sizes $\eta_{N,t}$ and $\eta_{T,t}$ are defined in \eqref{eq:adaptive_step_sizes}, while the Armijo scale $\gamma_t$ is selected according to \eqref{eq:armijo_condition}. CAT therefore preserves the host prior transition and adapts only its guidance correction.

In practice, CAT uses the learned score $\widehat s_t=s_\theta(x_t,t)$ to form
$\widehat\nu_t=\widehat s_t/\|\widehat s_t\|_2$ and directly decomposes the
guidance as $\widehat q_{N,t}=\widehat\nu_t(\widehat\nu_t^\top q_t)$ and
$\widehat q_{T,t}=q_t-\widehat q_{N,t}$.  CAT evaluates only the directional
curvature needed by the update.  For
$\widehat b_t=\|\widehat q_{T,t}\|_2>0$, define
\begin{equation}
\begin{aligned}
  u_t
  =
  \operatorname{stopgrad}
  \left(
    \frac{\widehat q_{T,t}}{\widehat b_t}
  \right),\qquad
  v_t
  =
  D_{x_t}
  \left(
    u_t^\top\widehat s_t
  \right),
  \qquad
  \widehat K_t
  =
  \frac{|u_t^\top v_t|}{\|\widehat s_t\|_2}.
\end{aligned}
  \label{eq:matrix_free_curvature}
\end{equation}
Here $D_{x_t}$ denotes reverse-mode differentiation with
respect to $x_t$.  Because $u_t$ is stop-gradient, this evaluates the required
quadratic form in one backward pass using only $D$-dimensional auxiliary
vectors.  If $\widehat b_t=0$, we set $\widehat K_t=0$ and skip the pass.

The complete CAT procedure decomposes the guidance gradient using the learned
score, evaluates \eqref{eq:matrix_free_curvature}, solves the scalar dual
equation, and calibrates the proposal by Armijo backtracking. Algorithm~\ref{alg:catdps} summarizes
the resulting implementation.

\begin{minipage}{\linewidth}
\begin{lstlisting}[caption={DPS+CAT.},label={alg:catdps}]
Input: state $x_{t+\Delta t}^{-}$, observation $y$,
       initial step $\alpha_t$, budget coefficient $\rho$,
       Armijo constants $c,\beta$
$x_t\leftarrow$ prior_step$(x_{t+\Delta t}^{-})$
$q_t\leftarrow\nabla L_t(x_t)$;  $\widehat s_t\leftarrow s_\theta(x_t,t)$
$\widehat\nu_t\leftarrow\widehat s_t/\|\widehat s_t\|_2$
$\widehat q_{N,t}\leftarrow
  \widehat\nu_t(\widehat\nu_t^\top q_t)$
$\widehat q_{T,t}\leftarrow q_t-\widehat q_{N,t}$
$\widehat a_t\leftarrow\|\widehat q_{N,t}\|_2$;
$\widehat b_t\leftarrow\|\widehat q_{T,t}\|_2$
if $\widehat b_t>0$ then
    $u_t\leftarrow
      \operatorname{stopgrad}(\widehat q_{T,t}/\widehat b_t)$
    $v_t\leftarrow
      \operatorname{autograd}_{x_t}(u_t^\top\widehat s_t)$
    $\widehat K_t\leftarrow|u_t^\top v_t|/\|\widehat s_t\|_2$
else
    $\widehat K_t\leftarrow0$
end if
$R_t\leftarrow\rho\sigma_t$
if $\alpha_t\widehat a_t+
   \frac12\widehat K_t(\alpha_t\widehat b_t)^2\leq R_t$ then
    $\lambda_t\leftarrow0$
else
    solve $\widehat F_t(\lambda_t)=R_t$ by bisection
end if
$r_{N,t}\leftarrow\alpha_t(\widehat a_t-\lambda_t)_+$
$r_{T,t}\leftarrow
  \alpha_t\widehat b_t/(1+\alpha_t\lambda_t\widehat K_t)$
$d_t\leftarrow
  -r_{N,t}\widehat q_{N,t}/\widehat a_t
  -r_{T,t}\widehat q_{T,t}/\widehat b_t$
$\gamma_t\leftarrow1$
while $d_t\neq0$ and
      $L_t(x_t+\gamma_td_t)>
      L_t(x_t)+c\gamma_tq_t^\top d_t$ do
    $\gamma_t\leftarrow\beta\gamma_t$
end while
return $x_t^{-}\leftarrow x_t+\gamma_td_t$
\end{lstlisting}
\end{minipage}

\section{Experiment}
\label{sec:experiment}

\subsection{Main Results}
\label{sec:main_results}

\paragraph{Evaluation protocol.}
We evaluate on 100 images from FFHQ and 100 images from ImageNet at the native resolution of $256\times256$. The benchmark covers random-mask and box inpainting, super-resolution, motion and Gaussian deblurring, high-dynamic-range (HDR) reconstruction, and nonlinear deblurring. Complete operator and measurement-noise settings are provided in Appendix~\ref{app:reproducibility}. We report PSNR, SSIM, LPIPS, and FID; higher PSNR and SSIM are better, whereas lower LPIPS and FID are better.

\paragraph{Methods and plugin protocol.}
We therefore attach CAT correction to multiple host samplers and report the resulting pairs separately: DPS~\cite{chung2023diffusion} and DPS+CAT in pixel space, and PSLD~\cite{rout2023solving}/PSLD+CAT and ReSample~\cite{song2024solving}/ReSample+CAT in latent space. This design tests whether the proposed geometric correction transfers across solver families. We additionally include DiffState-enhanced variants~\cite{zirvi2025diffusion} of the same hosts, together with the pixel-space baselines DPS, DDRM~\cite{kawar2022denoising}, DDNM~\cite{wang2023zeroshot}, DAPS~\cite{zhang2025improving}, DCDP~\cite{li2026dcdp}, and DiffRGD~\cite{liao2026diffrgd}. The algorithm hyperparameter settings used in these experiments are provided in Appendix~\ref{app:reproducibility}.

\begin{table}[p]
\centering
\caption{Main results on FFHQ and ImageNet using 100 images per dataset. Higher PSNR/SSIM and lower LPIPS/FID are better. \textbf{Bold} denotes the best result within each task and type.}
\label{tab:main_results}
\begingroup
\tiny
\setlength{\tabcolsep}{1.3pt}
\renewcommand{\arraystretch}{0.74}
\newcommand{\best}[1]{\textbf{#1}}
\newcommand{\second}[1]{#1}
\resizebox{\textwidth}{!}{%
\begin{tabular}{@{}>{\centering\arraybackslash}p{3.20cm}|c|>{\raggedright\arraybackslash}p{3.20cm}|*{4}{>{\centering\arraybackslash}p{0.95cm}}|*{4}{>{\centering\arraybackslash}p{0.95cm}}@{}}
  \toprule
  \textbf{Task} & \textbf{Type} & \textbf{Method} & \multicolumn{4}{c|}{\textbf{FFHQ}} & \multicolumn{4}{c}{\textbf{ImageNet}} \\
  \cmidrule(lr){4-7}\cmidrule(lr){8-11}
  & & & PSNR $\uparrow$ & SSIM $\uparrow$ & LPIPS $\downarrow$ & FID $\downarrow$ & PSNR $\uparrow$ & SSIM $\uparrow$ & LPIPS $\downarrow$ & FID $\downarrow$ \\
  \midrule

  \multirow{14}{3.20cm}{\centering Super-resolution $4\times$} & \multirow{8}{*}{\emph{Pixel}} & DAPS & 28.07 & 0.749 & 0.2773 & 76.15 & 25.54 & 0.642 & 0.3759 & 113.16 \\*
   &  & DCDP & 26.55 & 0.765 & 0.3074 & 103.17 & 25.80 & 0.681 & 0.3808 & 128.53 \\*
   &  & DDNM & \best{28.80} & \best{0.822} & 0.2435 & 80.38 & \best{26.84} & \best{0.743} & \second{0.3083} & 102.82 \\*
   &  & DDRM & \second{28.47} & \second{0.814} & 0.2511 & 84.87 & \second{26.43} & \second{0.730} & 0.3175 & \second{99.85} \\*
   &  & DiffRGD & 27.49 & 0.778 & \second{0.2266} & \second{64.60} & 23.96 & 0.625 & 0.3687 & 121.27 \\*
   &  & DPS & 24.01 & 0.672 & 0.3176 & 88.23 & 22.09 & 0.542 & 0.4486 & 166.28 \\*
   &  & DPS+Diffstate & 22.67 & 0.629 & 0.3475 & 90.20 & 20.85 & 0.494 & 0.4933 & 192.38 \\*
   & & DPS+CAT (ours) & 27.58 & 0.780 & \best{0.2132} & \best{54.63} & 25.67 & 0.691 & \best{0.2977} & \best{82.44} \\
  \cmidrule(lr){2-11}
   & \multirow{6}{*}{\emph{Latent}} & PSLD & 23.91 & 0.610 & 0.3973 & 134.91 & 22.62 & 0.531 & 0.4974 & 265.51 \\*
   &  & PSLD+DiffState & 24.53 & 0.649 & \second{0.3444} & \second{94.91} & 21.20 & 0.474 & 0.5525 & 299.54 \\*
   & & PSLD+CAT (ours) & \best{27.16} & \best{0.754} & \best{0.2497} & \best{71.01} & 23.43 & 0.575 & 0.4487 & 233.00 \\*
   &  & ReSample & 23.25 & 0.585 & 0.4635 & 164.56 & \second{24.42} & \second{0.637} & \second{0.3618} & \second{150.49} \\*
   &  & Resample+Diffstate & 21.31 & 0.436 & 0.5914 & 291.91 & 23.36 & 0.570 & 0.4467 & 235.14 \\*
   & & Resample+CAT (ours) & \second{26.80} & \second{0.720} & 0.3719 & 124.63 & \best{25.41} & \best{0.694} & \best{0.3538} & \best{138.28} \\
  \midrule

  \multirow{14}{3.20cm}{\centering Inpainting (box)} & \multirow{8}{*}{\emph{Pixel}} & DAPS & \second{24.46} & 0.752 & 0.2128 & 54.73 & \best{20.94} & 0.711 & 0.2754 & \second{97.58} \\*
   &  & DCDP & 22.87 & 0.750 & 0.3009 & 103.14 & 19.93 & 0.757 & 0.2694 & 122.64 \\*
   &  & DDNM & \best{24.51} & \best{0.860} & \second{0.1553} & 48.08 & \second{20.74} & \best{0.802} & \second{0.2172} & 99.90 \\*
   &  & DDRM & 22.18 & 0.818 & 0.2117 & 72.35 & 19.14 & 0.759 & 0.2685 & 121.00 \\*
   &  & DiffRGD & 23.33 & 0.839 & 0.1664 & \second{47.21} & 19.30 & 0.707 & 0.3359 & 154.77 \\*
   &  & DPS & 23.60 & 0.801 & 0.2201 & 65.92 & 19.97 & 0.695 & 0.3295 & 121.47 \\*
   &  & DPS+Diffstate & 23.23 & 0.736 & 0.2626 & 72.40 & 19.74 & 0.626 & 0.3818 & 137.10 \\*
   & & DPS+CAT (ours) & 24.35 & \second{0.849} & \best{0.1437} & \best{38.88} & 20.01 & \second{0.786} & \best{0.1769} & \best{64.99} \\
  \cmidrule(lr){2-11}
   & \multirow{6}{*}{\emph{Latent}} & PSLD & 23.50 & 0.857 & 0.1163 & 41.16 & 19.71 & 0.804 & 0.1956 & 120.25 \\*
   &  & PSLD+DiffState & 23.78 & 0.862 & 0.1140 & 40.54 & 19.74 & 0.804 & 0.2005 & 125.97 \\*
   & & PSLD+CAT (ours) & \best{24.25} & \second{0.876} & \best{0.0983} & \best{33.38} & 20.19 & 0.814 & 0.1888 & 115.24 \\*
   &  & ReSample & 20.63 & 0.838 & 0.1444 & 80.55 & \second{20.79} & \second{0.822} & \second{0.1769} & \second{108.18} \\*
   &  & Resample+Diffstate & 20.66 & 0.793 & 0.1919 & 141.88 & 19.86 & 0.811 & \best{0.1766} & 108.59 \\*
   & & Resample+CAT (ours) & \second{23.83} & \best{0.876} & \second{0.1079} & \second{40.42} & \best{21.10} & \best{0.827} & 0.1782 & \best{106.21} \\
  \midrule

  \multirow{14}{3.20cm}{\centering Inpainting (random)} & \multirow{8}{*}{\emph{Pixel}} & DAPS & 29.44 & 0.784 & 0.2144 & 52.17 & 27.35 & 0.725 & 0.2516 & \second{48.76} \\*
   &  & DCDP & 27.17 & 0.786 & 0.2838 & 98.72 & 26.67 & 0.743 & 0.3111 & 88.60 \\*
   &  & DDNM & 29.60 & 0.862 & 0.1845 & 59.98 & 27.32 & 0.785 & \second{0.2394} & 67.16 \\*
   &  & DDRM & 25.65 & 0.768 & 0.2791 & 109.90 & 23.24 & 0.633 & 0.4105 & 170.01 \\*
   &  & DiffRGD & \best{30.88} & \best{0.889} & \second{0.1329} & \second{32.88} & \second{28.71} & \second{0.822} & 0.2415 & 72.51 \\*
   &  & DPS & 28.08 & 0.804 & 0.2362 & 72.72 & 26.17 & 0.707 & 0.3328 & 105.54 \\*
   &  & DPS+Diffstate & 26.24 & 0.740 & 0.2768 & 80.76 & 24.41 & 0.633 & 0.3962 & 142.86 \\*
   & & DPS+CAT (ours) & \second{30.69} & \second{0.883} & \best{0.1260} & \best{29.56} & \best{29.22} & \best{0.841} & \best{0.1781} & \best{39.52} \\
  \cmidrule(lr){2-11}
   & \multirow{6}{*}{\emph{Latent}} & PSLD & 23.86 & 0.651 & 0.4148 & 148.66 & 24.93 & 0.699 & 0.4098 & 188.71 \\*
   &  & PSLD+DiffState & 22.64 & 0.600 & 0.4572 & 161.32 & 23.56 & 0.647 & 0.4622 & 245.96 \\*
   & & PSLD+CAT (ours) & \second{29.85} & \second{0.858} & 0.2117 & 69.04 & 25.31 & 0.722 & 0.3842 & 176.52 \\*
   &  & ReSample & 29.61 & 0.849 & \second{0.1911} & \second{55.54} & \second{27.38} & \second{0.802} & \second{0.2667} & \second{85.86} \\*
   &  & Resample+Diffstate & 29.14 & 0.833 & 0.2338 & 86.92 & 27.03 & 0.776 & 0.3091 & 110.79 \\*
   & & Resample+CAT (ours) & \best{30.23} & \best{0.866} & \best{0.1725} & \best{51.99} & \best{27.46} & \best{0.805} & \best{0.2650} & \best{85.22} \\
  \midrule

  \multirow{14}{3.20cm}{\centering Gaussian deblurring} & \multirow{8}{*}{\emph{Pixel}} & DAPS & 28.23 & 0.752 & 0.2687 & 72.33 & 25.95 & 0.662 & 0.3578 & 108.40 \\*
   &  & DCDP & 28.17 & 0.790 & 0.2650 & 81.75 & 25.89 & 0.673 & 0.4067 & 149.88 \\*
   &  & DDNM & \best{29.19} & \best{0.828} & 0.2707 & 95.79 & \second{26.77} & \second{0.733} & 0.3347 & 115.95 \\*
   &  & DDRM & \second{28.66} & \second{0.817} & 0.2499 & 83.63 & \best{26.82} & \best{0.735} & \second{0.3168} & \second{105.26} \\*
   &  & DiffRGD & 26.46 & 0.750 & \second{0.2259} & \second{64.63} & 22.62 & 0.568 & 0.4158 & 141.71 \\*
   &  & DPS & 25.15 & 0.698 & 0.2719 & 73.70 & 22.42 & 0.547 & 0.3834 & 119.95 \\*
   &  & DPS+Diffstate & 24.53 & 0.678 & 0.2890 & 76.71 & 22.50 & 0.547 & 0.3951 & 136.01 \\*
   & & DPS+CAT (ours) & 27.20 & 0.764 & \best{0.2064} & \best{50.69} & 24.86 & 0.657 & \best{0.2736} & \best{74.27} \\
  \cmidrule(lr){2-11}
   & \multirow{6}{*}{\emph{Latent}} & PSLD & 22.96 & 0.561 & 0.3875 & 119.22 & 20.76 & 0.464 & 0.5297 & 270.49 \\*
   &  & PSLD+DiffState & 22.51 & 0.549 & 0.3949 & 109.64 & 20.77 & 0.429 & 0.5379 & 277.55 \\*
   & & PSLD+CAT (ours) & 25.59 & \second{0.706} & \second{0.3141} & 97.68 & 22.01 & 0.519 & 0.5165 & 275.79 \\*
   &  & ReSample & \second{25.88} & 0.643 & 0.3274 & \second{94.59} & \second{24.65} & \second{0.640} & \second{0.4351} & \second{186.59} \\*
   &  & Resample+Diffstate & 24.45 & 0.565 & 0.3857 & 112.11 & 24.26 & 0.621 & 0.4542 & 203.70 \\*
   & & Resample+CAT (ours) & \best{27.91} & \best{0.779} & \best{0.2460} & \best{75.52} & \best{24.85} & \best{0.652} & \best{0.3981} & \best{164.85} \\
  \midrule

  \multirow{12}{3.20cm}{\centering Motion deblurring} & \multirow{6}{*}{\emph{Pixel}} & DAPS & \best{30.28} & 0.811 & 0.2093 & 56.99 & \second{28.62} & \second{0.768} & \second{0.2487} & \second{54.31} \\*
   &  & DCDP & 29.53 & 0.808 & 0.2185 & 57.84 & 27.27 & 0.704 & 0.2961 & 72.03 \\*
   &  & DiffRGD & 28.81 & \second{0.816} & \second{0.1913} & \second{53.84} & 23.32 & 0.597 & 0.4284 & 155.59 \\*
   &  & DPS & 26.31 & 0.740 & 0.2532 & 70.15 & 24.55 & 0.643 & 0.3479 & 102.60 \\*
   &  & DPS+Diffstate & 25.33 & 0.706 & 0.2773 & 75.78 & 23.55 & 0.597 & 0.3864 & 128.44 \\*
   & & DPS+CAT (ours) & \second{29.80} & \best{0.833} & \best{0.1634} & \best{39.25} & \best{28.76} & \best{0.800} & \best{0.2072} & \best{48.09} \\
  \cmidrule(lr){2-11}
   & \multirow{6}{*}{\emph{Latent}} & PSLD & 23.79 & 0.606 & 0.3706 & 117.02 & 21.68 & 0.488 & 0.5275 & 280.95 \\*
   &  & PSLD+DiffState & 23.40 & 0.601 & 0.3695 & 100.68 & 20.33 & 0.433 & 0.5752 & 303.47 \\*
   & & PSLD+CAT (ours) & 27.05 & \second{0.751} & 0.2670 & 76.61 & 22.77 & 0.558 & 0.4825 & 253.68 \\*
   &  & ReSample & \second{28.19} & 0.742 & \second{0.2589} & \second{74.53} & \second{25.34} & \second{0.675} & \second{0.4004} & \second{173.04} \\*
   &  & Resample+Diffstate & 28.04 & 0.746 & 0.2670 & 85.76 & 24.94 & 0.652 & 0.4246 & 190.48 \\*
   & & Resample+CAT (ours) & \best{29.23} & \best{0.806} & \best{0.2109} & \best{66.22} & \best{25.45} & \best{0.684} & \best{0.3597} & \best{140.46} \\
  \midrule

  \multirow{8}{3.20cm}{\centering HDR} & \multirow{5}{*}{\emph{Pixel}} & DAPS & \second{26.98} & \second{0.828} & \second{0.1977} & \second{53.10} & \best{26.12} & \second{0.827} & \second{0.2051} & \second{45.60} \\*
   &  & DCDP & \best{27.09} & \best{0.842} & 0.2222 & 61.53 & \second{26.11} & \best{0.831} & 0.2126 & 50.16 \\*
   &  & DPS & 23.59 & 0.721 & 0.2938 & 82.85 & 17.12 & 0.460 & 0.5375 & 184.88 \\*
   &  & DPS+Diffstate & 24.00 & 0.717 & 0.2931 & 78.95 & 20.34 & 0.525 & 0.5038 & 186.90 \\*
   & & DPS+CAT (ours) & 25.91 & 0.819 & \best{0.1664} & \best{42.80} & 25.93 & 0.822 & \best{0.1763} & \best{41.82} \\
  \cmidrule(lr){2-11}
   & \multirow{3}{*}{\emph{Latent}} & ReSample & 24.98 & \second{0.769} & \second{0.2877} & \second{91.27} & -- & -- & -- & -- \\*
   &  & Resample+Diffstate & \best{25.67} & 0.757 & 0.2927 & 94.89 & -- & -- & -- & -- \\*
   & & Resample+CAT (ours) & \second{25.61} & \best{0.794} & \best{0.2561} & \best{81.15} & -- & -- & -- & -- \\
  \midrule

  \multirow{8}{3.20cm}{\centering Nonlinear deblurring} & \multirow{5}{*}{\emph{Pixel}} & DAPS & \second{26.25} & \second{0.710} & \second{0.2809} & \second{80.99} & \best{25.16} & \second{0.661} & \second{0.3429} & \second{119.91} \\*
   &  & DCDP & 24.22 & 0.710 & 0.3182 & 101.08 & 22.17 & 0.531 & 0.4619 & 176.27 \\*
   &  & DPS & 21.97 & 0.627 & 0.3219 & 85.68 & 20.05 & 0.502 & 0.4320 & 169.87 \\*
   &  & DPS+Diffstate & 21.57 & 0.607 & 0.3336 & 83.03 & 18.78 & 0.428 & 0.5075 & 194.37 \\*
   & & DPS+CAT (ours) & \best{26.30} & \best{0.762} & \best{0.2197} & \best{55.17} & \second{24.95} & \best{0.690} & \best{0.2583} & \best{73.02} \\
  \cmidrule(lr){2-11}
   & \multirow{3}{*}{\emph{Latent}} & ReSample & \second{26.75} & \second{0.742} & \second{0.2790} & \second{81.67} & -- & -- & -- & -- \\*
   &  & Resample+Diffstate & 25.96 & 0.712 & 0.3191 & 108.60 & -- & -- & -- & -- \\*
   & & Resample+CAT (ours) & \best{27.00} & \best{0.750} & \best{0.2666} & \best{74.39} & -- & -- & -- & -- \\
\bottomrule
\end{tabular}%
}
\endgroup
\end{table}

\begin{figure}[!htbp]
  \centering
  \setlength{\tabcolsep}{0.4pt}
  \newlength{\mainqsize}
  \newcommand{\mainqimg}[1]{\includegraphics[width=0.191\linewidth]{#1}}
  \newcommand{\mainqredbox}[4]{%
    \put(#1,#2){\color{red}\linethickness{0.7pt}\framebox(#3,#4){}}%
  }
  \newcommand{\mainqannot}[2]{%
    \begingroup
    \setlength{\mainqsize}{0.191\linewidth}%
    \setlength{\unitlength}{\dimexpr\mainqsize/256\relax}%
    \begin{picture}(256,256)
      \put(0,0){\includegraphics[width=\mainqsize]{#1}}%
      #2%
    \end{picture}%
    \endgroup
  }
  \newcommand{\qualrowmarked}[3]{%
    \mainqannot{figures/reference__main_results_FFHQ__daps__#1__label__#2.png}{#3} &
    \mainqimg{figures/reference__main_results_FFHQ__daps__#1__input__#2.png} &
    \mainqannot{figures/reference__main_results_FFHQ__dps-tubular__#1__recon__#2.png}{#3} &
    \mainqannot{figures/reference__main_results_FFHQ__dps-diffstate__#1__recon__#2.png}{#3} &
    \mainqannot{figures/reference__main_results_FFHQ__dps__#1__recon__#2.png}{#3}%
  }
  \newcommand{\qualpanel}[5]{%
    \begin{minipage}[t]{0.485\linewidth}
      \centering
      \begin{tabular}{@{}ccccc@{}}
        \scriptsize Original & \scriptsize Measurement & \scriptsize DPS+CAT &
        \fontsize{5.2}{6.0}\selectfont DPS+Diffstate & \scriptsize DPS \\
        \mainqimg{figures/reference__main_results_FFHQ__daps__#1__label__#2.png} &
        \mainqimg{figures/reference__main_results_FFHQ__daps__#1__input__#2.png} &
        \mainqimg{figures/reference__main_results_FFHQ__dps-tubular__#1__recon__#2.png} &
        \mainqimg{figures/reference__main_results_FFHQ__dps-diffstate__#1__recon__#2.png} &
        \mainqimg{figures/reference__main_results_FFHQ__dps__#1__recon__#2.png} \\
        \mainqimg{figures/reference__main_results_FFHQ__daps__#1__label__#3.png} &
        \mainqimg{figures/reference__main_results_FFHQ__daps__#1__input__#3.png} &
        \mainqimg{figures/reference__main_results_FFHQ__dps-tubular__#1__recon__#3.png} &
        \mainqimg{figures/reference__main_results_FFHQ__dps-diffstate__#1__recon__#3.png} &
        \mainqimg{figures/reference__main_results_FFHQ__dps__#1__recon__#3.png}
      \end{tabular}
      \vspace{0.5mm}

      \small\textbf{(#4)} #5
    \end{minipage}%
  }
  \newcommand{\qualpanelmarked}[7]{%
    \begin{minipage}[t]{0.485\linewidth}
      \centering
      \begin{tabular}{@{}ccccc@{}}
        \scriptsize Original & \scriptsize Measurement & \scriptsize DPS+CAT &
        \fontsize{5.2}{6.0}\selectfont DPS+Diffstate & \scriptsize DPS \\
        \qualrowmarked{#1}{#2}{#4} \\
        \qualrowmarked{#1}{#3}{#5}
      \end{tabular}
      \vspace{0.5mm}

      \small\textbf{(#6)} #7
    \end{minipage}%
  }

  \qualpanelmarked{inpainting_random}{00007}{00013}%
    {\mainqredbox{132}{73}{24}{24}}%
    {\mainqredbox{3}{3}{34}{31}}%
    {a}{Random inpainting.}\hfill
  \qualpanelmarked{motion_deblur}{00016}{00012}%
    {\mainqredbox{17}{16}{45}{44}}%
    {\mainqredbox{140}{151}{43}{32}}%
    {b}{Motion deblurring.}

  \vspace{2mm}

  \qualpanelmarked{gaussian_deblur}{00015}{00004}%
    {\mainqredbox{74}{122}{50}{35}\mainqredbox{99}{46}{61}{31}}%
    {\mainqredbox{198}{83}{29}{35}}%
    {c}{Gaussian deblurring.}\hfill
  \qualpanelmarked{super_resolution}{00019}{00009}%
    {\mainqredbox{83}{200}{86}{38}}%
    {\mainqredbox{99}{45}{61}{35}}%
    {d}{Super-resolution ($\times 4$).}

  \vspace{2mm}

  \qualpanelmarked{nonlinear_deblur}{00006}{00002}%
    {\mainqredbox{67}{68}{22}{42}\mainqredbox{175}{60}{24}{42}}%
    {\mainqredbox{68}{93}{32}{31}\mainqredbox{160}{99}{51}{35}}%
    {e}{Nonlinear deblurring.}\hfill
  \qualpanel{high_dynamic_range}{00011}{00018}{f}{HDR reconstruction.}

  \caption{Qualitative comparison of DPS-based methods on FFHQ across six representative inverse problems. Each panel shows two examples with the same original image and measurement. DPS+CAT denotes DPS equipped with CAT, while DPS+Diffstate is the corresponding DiffState baseline.}
  \label{fig:ffhq_main_qualitative}
\end{figure}
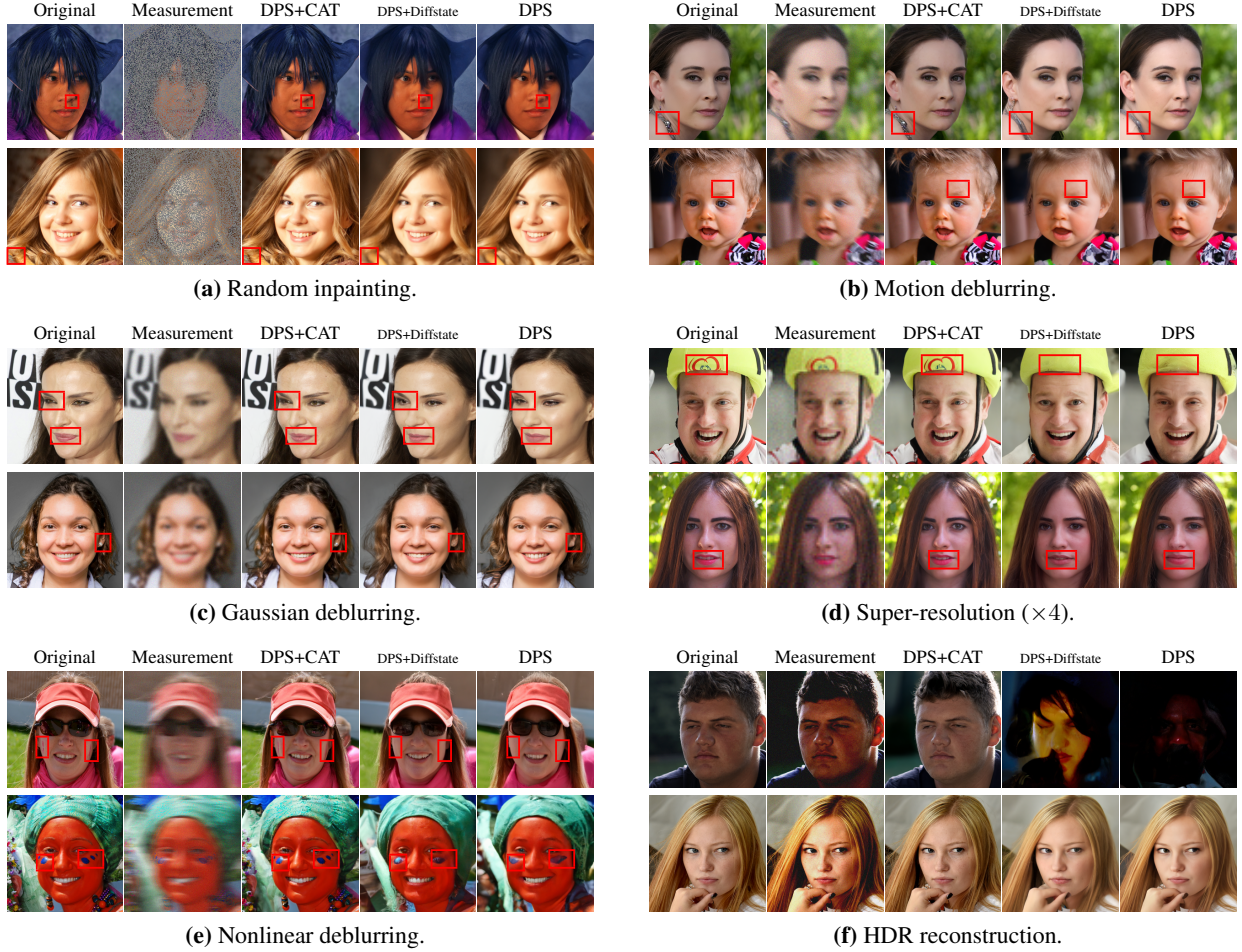

\paragraph{Quantitative comparison.}
Table~\ref{tab:main_results} summarizes all seven tasks on both FFHQ and ImageNet. On FFHQ, integrating CAT consistently improves the underlying solvers across all seven inverse problems, with particularly strong reductions in LPIPS and FID, indicating better perceptual fidelity. These gains extend to ImageNet, where DPS+CAT achieves the strongest perceptual results among the pixel-space methods across all reported tasks. Resample+CAT delivers the strongest overall latent-space results on the five linear tasks. The consistent improvements across DPS, PSLD, and ReSample demonstrate that CAT serves as an effective plug-and-play geometric correction rather than a solver-specific alternative. Figure~\ref{fig:ffhq_main_qualitative} further confirms these gains: across six representative tasks, DPS+CAT better preserves facial structure and fine details while suppressing local artifacts.

\FloatBarrier

\subsection{Robustness Analysis}
\label{sec:robustness_results}

We evaluate the robustness of CAT along three sources of sensitivity: the host guidance step size, the tubular radius, and the measurement-noise level. All sweeps use FFHQ motion deblurring and aggregate 100 images at each setting.

\begin{figure}[!htbp]
  \centering
  \includegraphics[width=\linewidth]{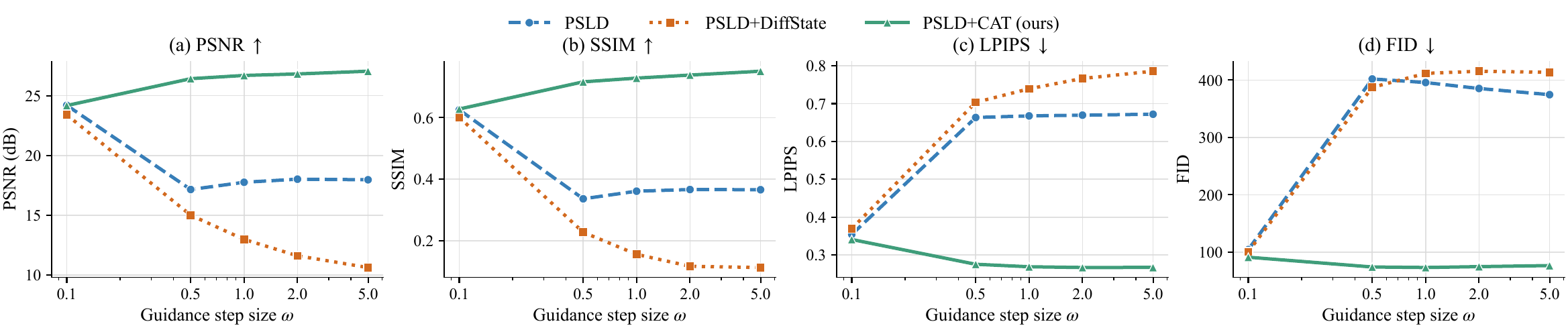}
  \caption{Robustness to the guidance step size $\omega$ on FFHQ motion deblurring. Each point aggregates 100 images.}
  \label{fig:robustness_step}
\end{figure}

\paragraph{Robustness to the guidance step size.}
The guidance step size controls the strength of the measurement-consistency updates applied by the host sampler. Although stronger guidance can improve data fidelity, an excessively large step may push the iterate away from regions well modeled by the learned prior. To assess this sensitivity, we sweep the PSLD guidance weight $\omega\in\{0.1,0.5,1,2,5\}$ on FFHQ motion deblurring, holding the latent data-consistency weight fixed at $\gamma=0.01$ and the measurement-noise level fixed at $\sigma_y=0.05$. All other hyperparameters remain unchanged. We evaluate PSLD, PSLD+DiffState, and PSLD+CAT on the same set of 100 images; Figure~\ref{fig:robustness_step} reports PSNR, SSIM, LPIPS, and FID across the sweep.

\begin{figure}[!htbp]
  \centering
  \setlength{\tabcolsep}{0.35pt}
  \newcommand{\stepqimg}[1]{\includegraphics[width=0.125\linewidth]{#1}}
  \newcommand{\stepqlabel}[1]{%
    \parbox[b][0.125\linewidth][c]{2.6mm}{%
      \centering\rotatebox[origin=c]{90}{\scriptsize #1}}}
  \begin{tabular}{@{}c@{\hspace{0.8mm}}ccccccc@{}}
    & \scriptsize Label & \scriptsize Measurement & \scriptsize $\omega=0.1$
    & \scriptsize $\omega=0.5$ & \scriptsize $\omega=1$
    & \scriptsize $\omega=2$ & \scriptsize $\omega=5$ \\[-0.3mm]
    \stepqlabel{PSLD}
    & \stepqimg{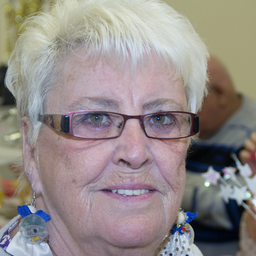}
    & \stepqimg{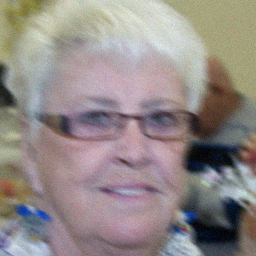}
    & \stepqimg{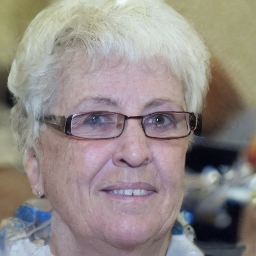}
    & \stepqimg{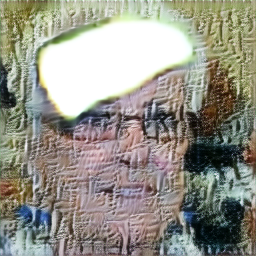}
    & \stepqimg{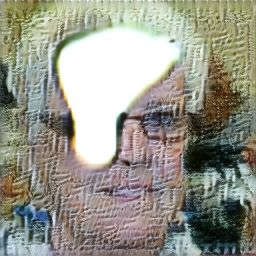}
    & \stepqimg{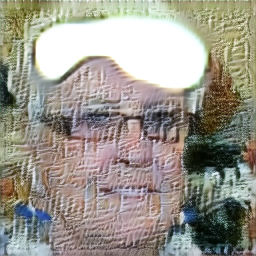}
    & \stepqimg{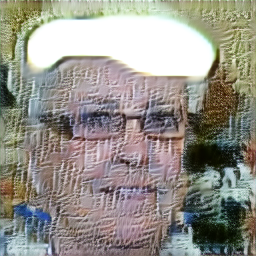} \\[-0.5mm]
    \stepqlabel{PSLD+DiffState}
    & \stepqimg{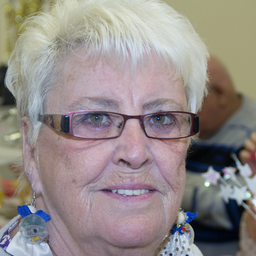}
    & \stepqimg{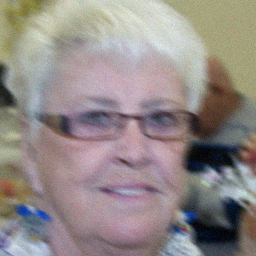}
    & \stepqimg{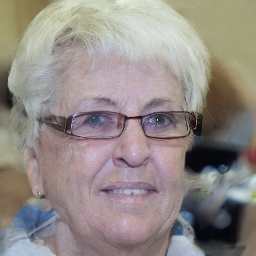}
    & \stepqimg{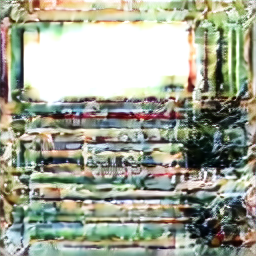}
    & \stepqimg{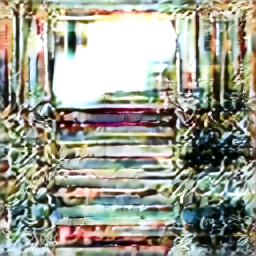}
    & \stepqimg{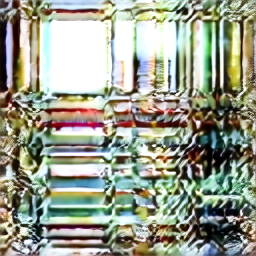}
    & \stepqimg{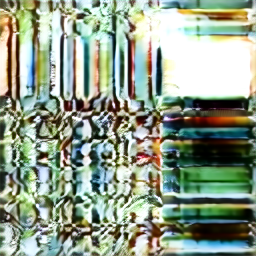} \\[-0.5mm]
    \stepqlabel{PSLD+CAT}
    & \stepqimg{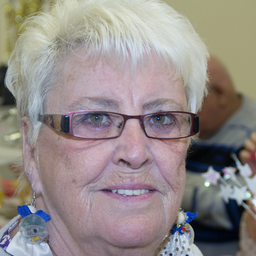}
    & \stepqimg{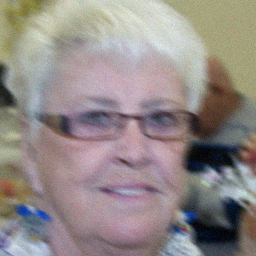}
    & \stepqimg{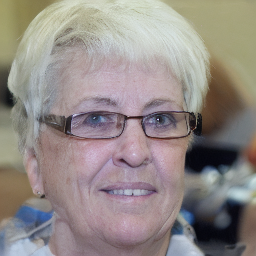}
    & \stepqimg{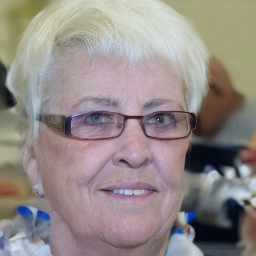}
    & \stepqimg{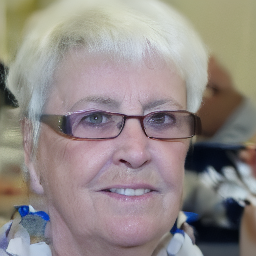}
    & \stepqimg{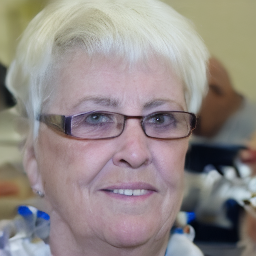}
    & \stepqimg{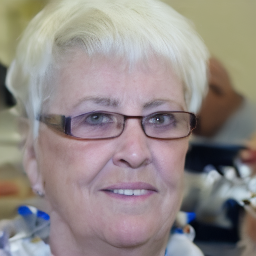}
  \end{tabular}
  \caption{Qualitative robustness to the guidance step size on FFHQ motion deblurring. Rows correspond to PSLD, PSLD+DiffState, and PSLD+CAT; columns show the ground-truth label, the blurred measurement, and reconstructions obtained with five values of $\omega$. All entries use the same input image.}
  \label{fig:robustness_step_qualitative}
\end{figure}

The quantitative and qualitative results consistently show that CAT substantially reduces sensitivity to the guidance step size. As the guidance becomes more aggressive, PSLD and PSLD+DiffState degrade sharply across the evaluation metrics, whereas PSLD+CAT maintains strong reconstruction quality over a broad operating range. Figure~\ref{fig:robustness_step_qualitative} further reveals pronounced structural distortions and color artifacts in the comparison methods at large step sizes, while PSLD+CAT preserves coherent facial geometry and a natural appearance. 

\FloatBarrier

\begin{figure}[!htbp]
  \centering
  \includegraphics[width=\linewidth]{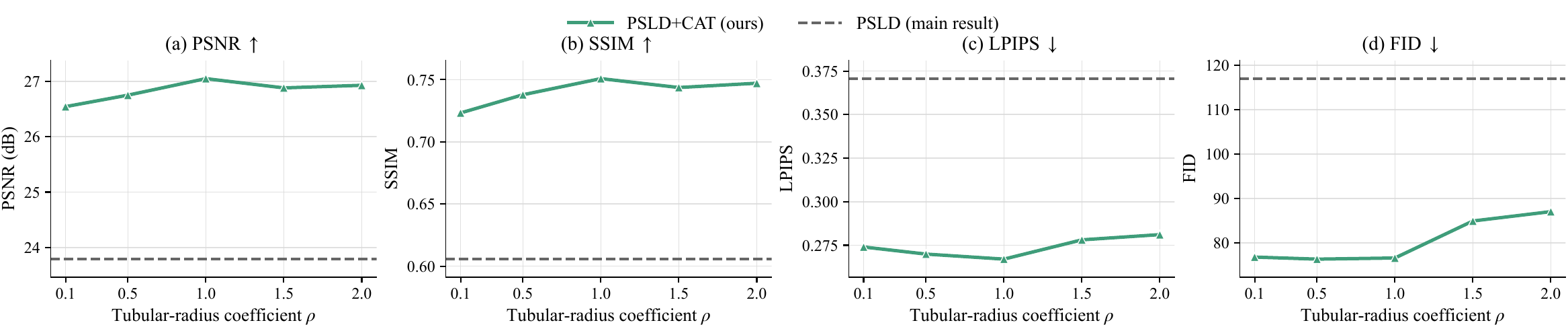}
  \caption{Robustness of PSLD+CAT to the tubular-radius coefficient $\rho$ on FFHQ motion deblurring. Each point aggregates 100 images, and the horizontal dashed line marks the PSLD result reported in the main experiment.}
  \label{fig:robustness_radius}
\end{figure}

\paragraph{Robustness to the tubular radius.}
The coefficient $\rho$ controls the noise-dependent tubular budget $R_t=\rho\sigma_t$: a small value restricts the correction to a narrow neighborhood of the local prior geometry, whereas a larger value permits more measurement-driven motion. We sweep $\rho\in\{0.1,0.5,1,1.5,2\}$ for PSLD+CAT on FFHQ motion deblurring while retaining the fixed host settings and measurement-noise level $\sigma_y=0.05$. Every radius is evaluated on the same 100 images. Figure~\ref{fig:robustness_radius} reports the resulting metric curves.

\begin{figure}[!htbp]
  \centering
  \setlength{\tabcolsep}{0.25pt}
  \newcommand{\radiusqimg}[1]{\includegraphics[width=0.115\linewidth]{#1}}
  \begin{tabular}{@{}cccccccc@{}}
    \scriptsize Label & \scriptsize Measurement
    & \fontsize{6.6}{7.2}\selectfont\shortstack{PSLD} & \scriptsize $\rho=0.1$
    & \scriptsize $\rho=0.5$ & \scriptsize $\rho=1$
    & \scriptsize $\rho=1.5$ & \scriptsize $\rho=2$ \\[-0.3mm]
    \radiusqimg{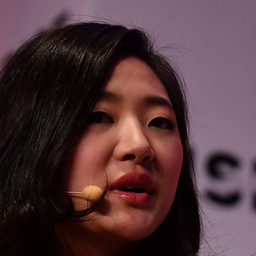}
    & \radiusqimg{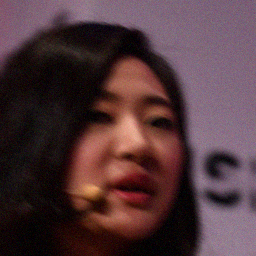}
    & \radiusqimg{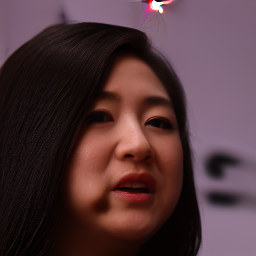}
    & \radiusqimg{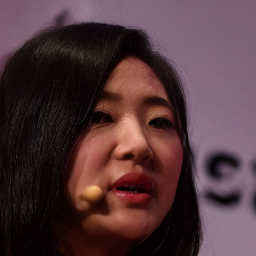}
    & \radiusqimg{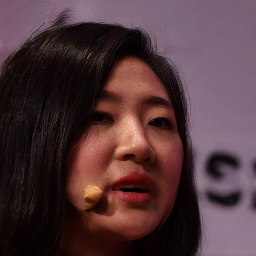}
    & \radiusqimg{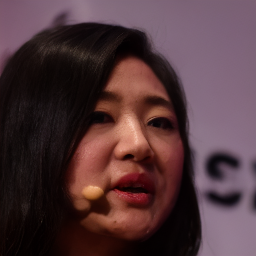}
    & \radiusqimg{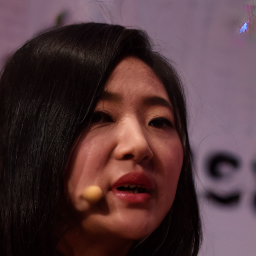}
    & \radiusqimg{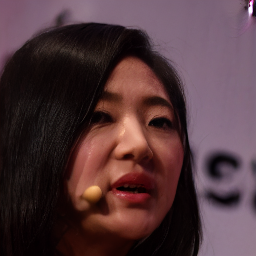}
  \end{tabular}
  \caption{Qualitative robustness of PSLD+CAT to the tubular-radius coefficient on FFHQ motion deblurring. Columns show the ground-truth label, the blurred measurement, the PSLD reconstruction from the main experiment, and PSLD+CAT reconstructions obtained with five values of $\rho$ for the same input.}
  \label{fig:robustness_radius_qualitative}
\end{figure}

The radius sweep confirms that PSLD+CAT is insensitive to the choice of $\rho$ over a broad operating range. Across all tested radii, PSLD+CAT consistently outperforms the PSLD result from the main experiment, shown by the horizontal dashed line in Figure~\ref{fig:robustness_radius}. The distortion metrics remain stable across the sweep, while Figure~\ref{fig:robustness_radius_qualitative} shows that PSLD+CAT preserves global facial structure more faithfully than PSLD across all tested radii. Overall, PSLD+CAT maintains reliable reconstruction quality without careful tuning of $\rho$, demonstrating robust behavior across a practical range of tubular radii.

\begin{figure}[!htbp]
  \centering
  \includegraphics[width=\linewidth]{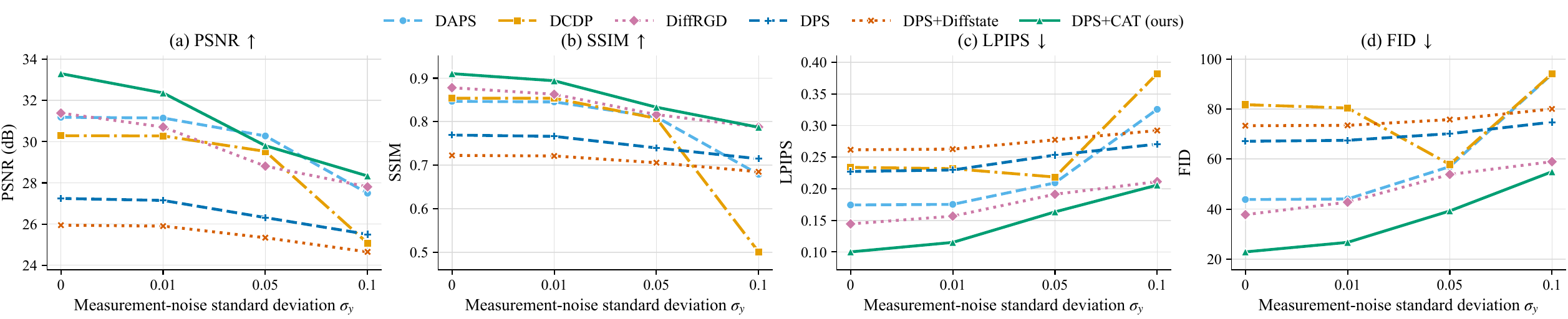}
  \caption{Robustness to the measurement-noise standard deviation $\sigma_y$ on FFHQ motion deblurring. Each point aggregates 100 images.}
  \label{fig:robustness_noise}
\end{figure}

\paragraph{Robustness to measurement noise.}
Measurement noise makes data-consistency guidance less reliable by perturbing the observations used throughout posterior sampling. We evaluate this sensitivity by varying $\sigma_y\in\{0,0.01,0.05,0.1\}$ while fixing the motion kernel and all solver hyperparameters. We compare DAPS, DCDP, DiffRGD, DPS, DPS+Diffstate, and DPS+CAT on the same 100 FFHQ images. Figure~\ref{fig:robustness_noise} reports one curve per method for each metric.

\begin{figure}[!htbp]
  \centering
  \setlength{\tabcolsep}{0.25pt}
  \newcommand{\noiseqimg}[1]{\includegraphics[width=0.11\linewidth]{#1}}
  \newcommand{\noiseqblank}{\rule{0.11\linewidth}{0pt}}
  \newcommand{\noiseqsigma}[1]{%
    \parbox[b][0.11\linewidth][c]{3.4mm}{%
      \centering\rotatebox[origin=c]{90}{\scriptsize $#1$}}}
  \begin{tabular}{@{}c@{\hspace{1.5mm}}c@{\hspace{0.8mm}}ccccccc@{}}
    \scriptsize Label &
    & \scriptsize Measurement & \scriptsize DAPS & \scriptsize DCDP
    & \scriptsize DiffRGD & \scriptsize DPS
    & \scriptsize\shortstack{DPS+\\Diffstate} & \scriptsize DPS+CAT \\[-0.3mm]
    \noiseqimg{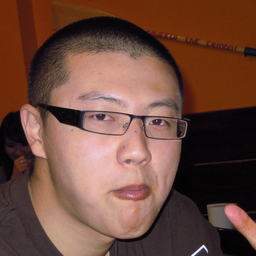}
    & \noiseqsigma{\sigma_y=0}
    & \noiseqimg{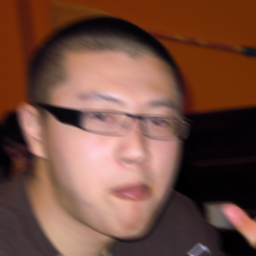}
    & \noiseqimg{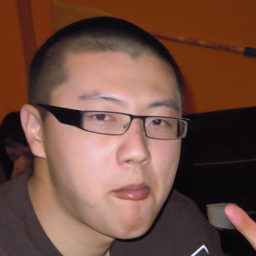}
    & \noiseqimg{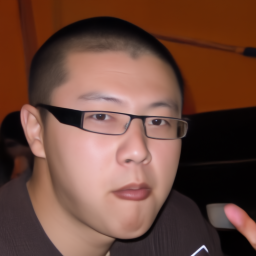}
    & \noiseqimg{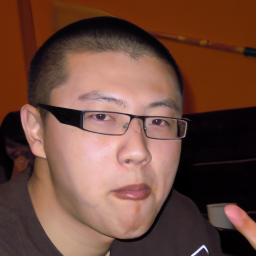}
    & \noiseqimg{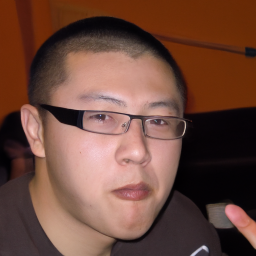}
    & \noiseqimg{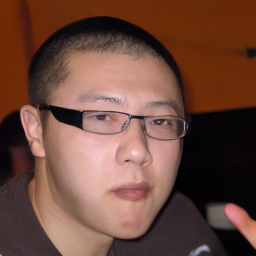}
    & \noiseqimg{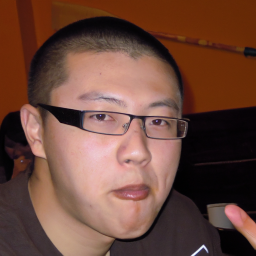}
    \\[-0.5mm]
    \noiseqblank
    & \noiseqsigma{\sigma_y=0.01}
    & \noiseqimg{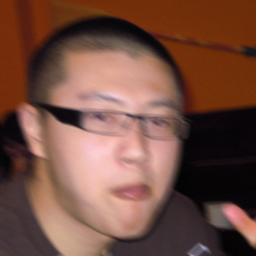}
    & \noiseqimg{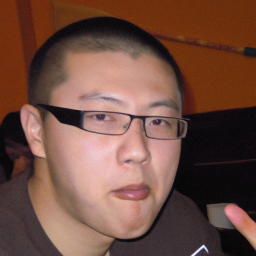}
    & \noiseqimg{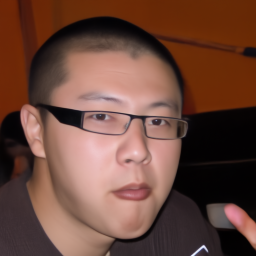}
    & \noiseqimg{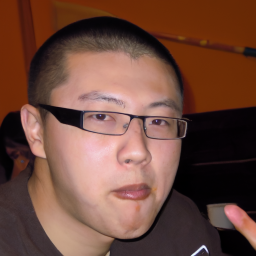}
    & \noiseqimg{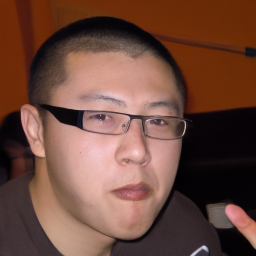}
    & \noiseqimg{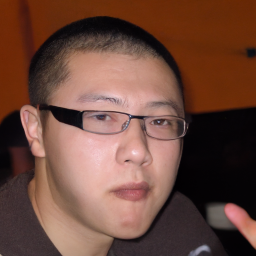}
    & \noiseqimg{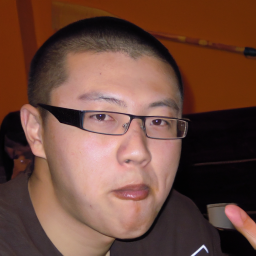}
    \\[-0.5mm]
    \noiseqblank
    & \noiseqsigma{\sigma_y=0.05}
    & \noiseqimg{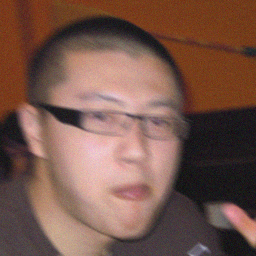}
    & \noiseqimg{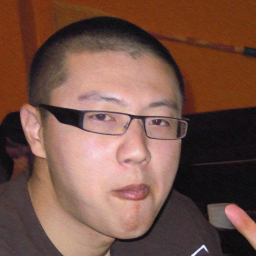}
    & \noiseqimg{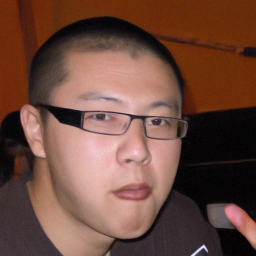}
    & \noiseqimg{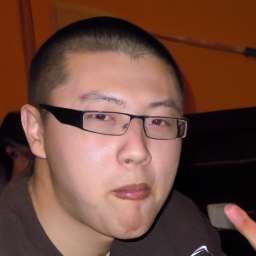}
    & \noiseqimg{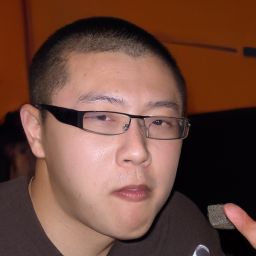}
    & \noiseqimg{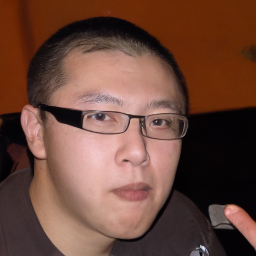}
    & \noiseqimg{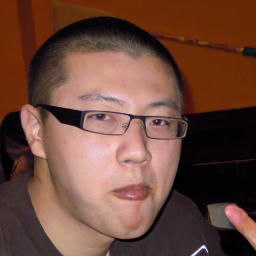}
    \\[-0.5mm]
    \noiseqblank
    & \noiseqsigma{\sigma_y=0.1}
    & \noiseqimg{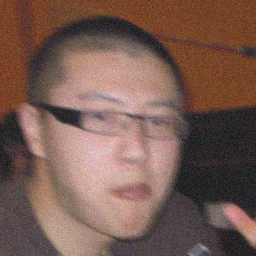}
    & \noiseqimg{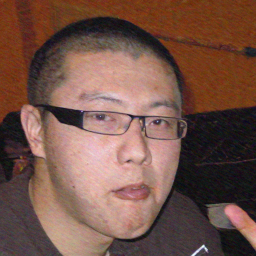}
    & \noiseqimg{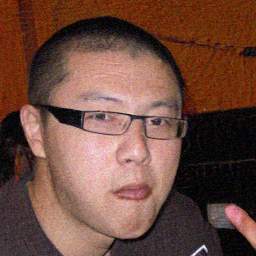}
    & \noiseqimg{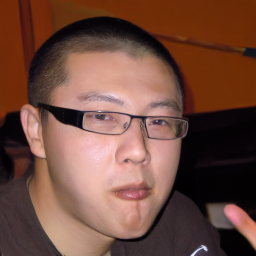}
    & \noiseqimg{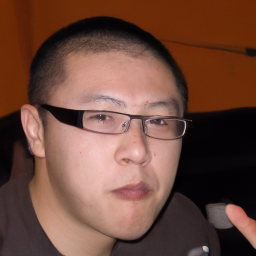}
    & \noiseqimg{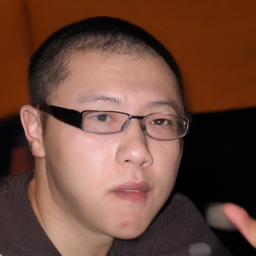}
    & \noiseqimg{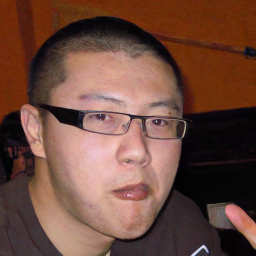}
  \end{tabular}
  \caption{Qualitative robustness to measurement noise on FFHQ motion deblurring. The left image shows the ground-truth label. Rows correspond to the four noise levels, while columns show the measurement and reconstructions from each method for the same input image.}
  \label{fig:robustness_noise_qualitative}
\end{figure}

The noise sweep demonstrates the strong robustness of DPS+CAT to measurement corruption. As the observations become increasingly unreliable, DPS+CAT consistently delivers the best perceptual quality while retaining highly competitive distortion metrics, indicating that CAT remains effective even under substantial measurement perturbations. Figure~\ref{fig:robustness_noise_qualitative} reinforces this advantage: DPS+CAT faithfully preserves facial boundaries and high-contrast details such as eyeglass frames, while avoiding the noise amplification and oversmoothing observed in several baselines. Together, these results show that CAT provides stable, high-quality reconstructions across a broad range of noise levels without relying on narrowly tuned measurement conditions.

\FloatBarrier

\subsection{Ablation Studies}
\label{sec:ablation_results}

We evaluate six diagnostic variants of CAT on FFHQ, using DPS for nonlinear deblurring and PSLD for motion deblurring.  Host applies strong guidance without CAT, whereas Host+Armijo adds only CAT's backtracking line search. First-order CAT disables the curvature term by setting $K_t=0$, Tangent-only removes the normal guidance component, and CAT w/o Armijo disables backtracking by fixing $\gamma_t=1$. Full CAT retains both the second-order tubular constraint and adaptive line search. The measurement-noise level is fixed at $\sigma_y=0.05$ throughout.

\begin{table}[!htbp]
  \centering
  \caption{Ablation of CAT on FFHQ using DPS for nonlinear deblurring and PSLD for motion deblurring. Each host and variant are evaluated on 100 images. ``1st'' and ``2nd'' denote first- and second-order tube models, respectively. Best results are shown in bold.}
  \label{tab:component_ablation}
  \setlength{\tabcolsep}{3.2pt}
  \renewcommand{\arraystretch}{1.08}
  \resizebox{\linewidth}{!}{%
  \begin{tabular}{@{}lcccccccccccc@{}}
    \toprule
    \textbf{Variant} & \textbf{N/T} & \textbf{Curv.} & \textbf{Tube} & \textbf{Armijo}
    & \multicolumn{4}{c}{\textbf{DPS: Nonlinear deblurring}}
    & \multicolumn{4}{c}{\textbf{PSLD: Motion deblurring}} \\
    \cmidrule(lr){6-9}\cmidrule(lr){10-13}
    & & & &
    & PSNR $\uparrow$ & SSIM $\uparrow$ & LPIPS $\downarrow$ & FID $\downarrow$
    & PSNR $\uparrow$ & SSIM $\uparrow$ & LPIPS $\downarrow$ & FID $\downarrow$ \\
    \midrule
    Host
    & No & No & -- & No
    & 17.06 & 0.326 & 0.5698 & 214.46
    & 17.96 & 0.366 & 0.6719 & 374.57 \\
    Host + Armijo
    & No & No & -- & Yes
    & 24.08 & 0.696 & 0.2751 & 71.71
    & 25.74 & 0.695 & 0.3482 & 123.15 \\
    First-order CAT
    & Yes & No & 1st & Yes
    & 24.26 & 0.700 & 0.2718 & 73.44
    & 25.80 & 0.699 & 0.3471 & 121.86 \\
    Tangent-only
    & Yes & No & Projection & Yes
    & 24.15 & 0.696 & 0.2731 & 75.45
    & 25.61 & 0.688 & 0.3504 & 125.81 \\
    CAT w/o Armijo
    & Yes & Yes & 2nd & No
    & 17.33 & 0.409 & 0.4555 & 192.73
    & 19.18 & 0.413 & 0.6243 & 323.10 \\
    \textbf{Full CAT}
    & Yes & Yes & 2nd & Yes
    & \textbf{26.30} & \textbf{0.762} & \textbf{0.2197} & \textbf{55.17}
    & \textbf{27.05} & \textbf{0.751} & \textbf{0.2670} & \textbf{76.61} \\
    \bottomrule
  \end{tabular}}
\end{table}

\begin{figure}[!htbp]
  \centering
  \setlength{\tabcolsep}{0.25pt}
  \newcommand{\ablationqimg}[1]{\includegraphics[width=0.112\linewidth]{#1}}
  \newcommand{\ablationqlabel}[1]{%
    \parbox[b][0.112\linewidth][c]{2.6mm}{%
      \centering\rotatebox[origin=c]{90}{\scriptsize #1}}}
  \begin{tabular}{@{}c@{\hspace{0.7mm}}cccccccc@{}}
    & \scriptsize Label & \scriptsize Measurement & \scriptsize Host
    & \scriptsize \shortstack{Host +\\Armijo}
    & \scriptsize \shortstack{First-order\\CAT}
    & \scriptsize \shortstack{Tangent-\\only}
    & \scriptsize \shortstack{CAT w/o\\Armijo}
    & \scriptsize Full CAT \\[-0.3mm]
    \ablationqlabel{DPS}
    & \ablationqimg{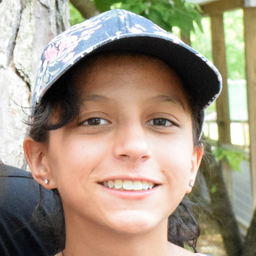}
    & \ablationqimg{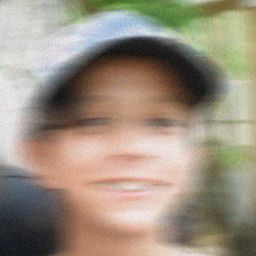}
    & \ablationqimg{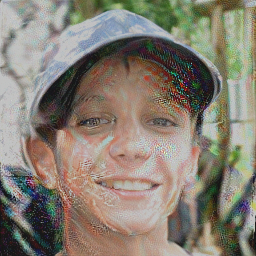}
    & \ablationqimg{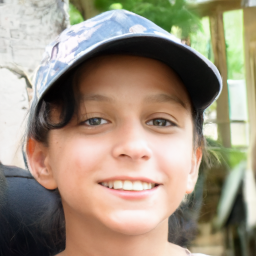}
    & \ablationqimg{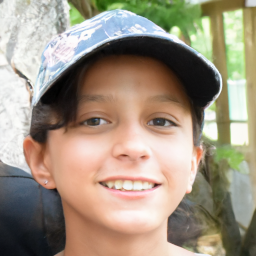}
    & \ablationqimg{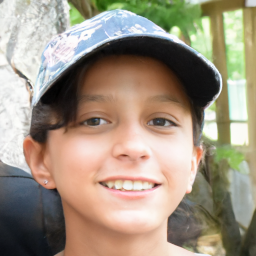}
    & \ablationqimg{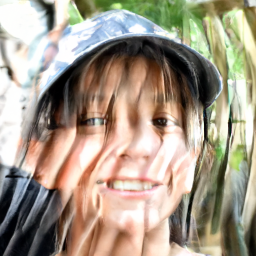}
    & \ablationqimg{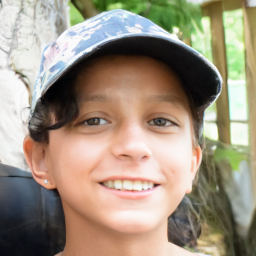} \\[-0.5mm]
    \ablationqlabel{PSLD}
    & \ablationqimg{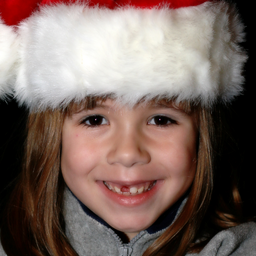}
    & \ablationqimg{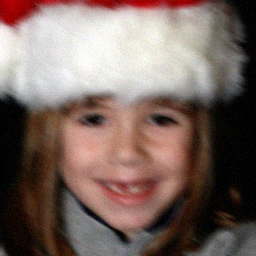}
    & \ablationqimg{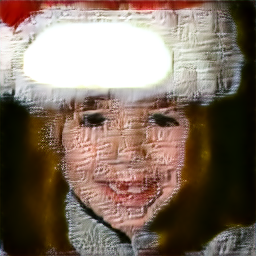}
    & \ablationqimg{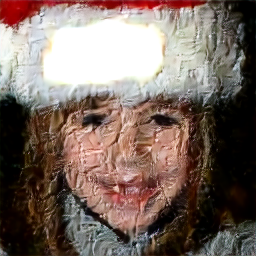}
    & \ablationqimg{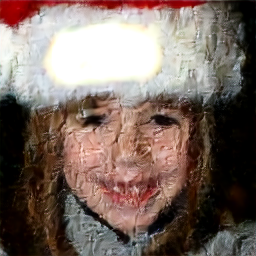}
    & \ablationqimg{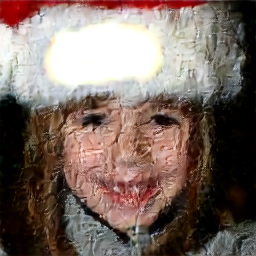}
    & \ablationqimg{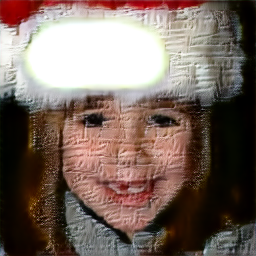}
    & \ablationqimg{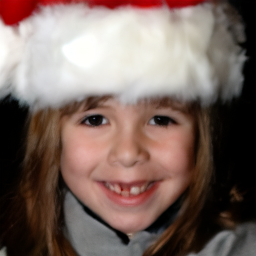}
  \end{tabular}
  \caption{Qualitative component ablation on FFHQ. The first row uses DPS for nonlinear deblurring, and the second uses PSLD for motion deblurring. Each row shares the same ground truth and measurement across variants.}
  \label{fig:component_ablation_qualitative}
\end{figure}

\paragraph{Component ablation.} Armijo backtracking substantially improves both host solvers, but does not explain the full gain of CAT. Full CAT achieves the best overall performance for both hosts, with consistent and non-negligible improvements over Host+Armijo. Conversely, CAT w/o Armijo provides only partial gains, showing that geometric control and finite-step loss calibration play complementary roles. Figure~\ref{fig:component_ablation_qualitative} supports this conclusion: Full CAT suppresses the severe artifacts remaining in the partial variants and recovers coherent facial structure.

\paragraph{Role of second-order geometry.} First-order CAT remains close to Host+Armijo, whereas Full CAT consistently performs better across both hosts. This gap indicates that constraining only first-order normal displacement is insufficient and that modeling curvature-induced tangent departure is important in practice. Tangent-only also trails Full CAT, suggesting that exact tangent projection discards useful measurement motion, while CAT benefits from retaining a controlled normal component.

\subsection{Efficiency Analysis}
\label{sec:efficiency_results}

We analyze the computational trade-offs of CAT on FFHQ nonlinear deblurring with DPS, using the same 100 images as in the main experiment. All runs are conducted on a single NVIDIA GeForce RTX 4090 GPU. We vary two factors: the number of reverse sampling steps $N$ and the Armijo period $p$ introduced above. All other method settings remain fixed within each sweep. Table~\ref{tab:efficiency_results} reports reconstruction quality, GPU memory usage, and wall-clock time.

\begin{table}[!htbp]
  \centering
  \caption{Efficiency of DPS+CAT on FFHQ nonlinear deblurring over 100 images using a single NVIDIA GeForce RTX 4090 GPU. Both blocks include the same DPS baseline for reference. The upper block varies the number of reverse sampling steps $N$ with Armijo applied at every step, while the lower block fixes $N=1000$ and varies the Armijo period $p$. Memory and time denote the recorded GPU memory usage and wall-clock runtime, respectively. Bold marks the best reconstruction quality within each block.}
  \label{tab:efficiency_results}
  \begingroup
  \small
  \setlength{\tabcolsep}{4.2pt}
  \renewcommand{\arraystretch}{1.05}
  \begin{tabular}{@{}llrrrrrr@{}}
    \toprule
    \textbf{Study} & \textbf{Setting} & \textbf{PSNR $\uparrow$} & \textbf{SSIM $\uparrow$} & \textbf{LPIPS $\downarrow$} & \textbf{FID $\downarrow$} & \textbf{Memory (MiB)} & \textbf{Time (s)} \\
    \midrule
    \multirow{6}{*}{Sampling steps}
      & DPS & 21.97 & 0.6269 & 0.3219 & 85.68 & 3490 & 38 \\
      & $N=200$  & 23.50 & 0.6851 & 0.2922 & 82.91 & 3610 & 26 \\
      & $N=400$  & 24.62 & 0.7152 & 0.2633 & 70.54 & 3610 & 51 \\
      & $N=600$  & 25.42 & 0.7380 & 0.2432 & 64.24 & 3610 & 78 \\
      & $N=800$  & 25.92 & 0.7525 & 0.2297 & 58.73 & 3612 & 103 \\
      & $N=1000$ & \textbf{26.30} & \textbf{0.7623} & \textbf{0.2197} & \textbf{55.17} & 3612 & 132 \\
    \midrule
    \multirow{5}{*}{Armijo period}
      & DPS & 21.97 & 0.6269 & 0.3219 & 85.68 & 3490 & 38 \\
      & $p=5$ & 24.85 & 0.7292 & 0.2473 & 67.04 & 3616 & 89 \\
      & $p=3$ & 24.82 & 0.7285 & 0.2472 & 62.92 & 3614 & 97 \\
      & $p=2$ & 24.48 & 0.7161 & 0.2561 & 69.29 & 3614 & 106 \\
      & $p=1$ (Full CAT) & \textbf{26.30} & \textbf{0.7623} & \textbf{0.2197} & \textbf{55.17} & 3612 & 132 \\
    \bottomrule
  \end{tabular}
  \endgroup
\end{table}

\paragraph{Armijo frequency.}
The Armijo period provides a flexible efficiency--quality trade-off: less frequent searches reduce the runtime overhead while preserving a clear quality advantage over DPS. When reconstruction quality is the priority, performing backtracking at every step yields the strongest and most consistent performance across the evaluation metrics. Overall, CAT provides a practical and adaptable enhancement whose computational cost can be adjusted without forfeiting its core reconstruction benefits.

\paragraph{Number of sampling steps.}
Compared with DPS, DPS+CAT improves all four reconstruction metrics at every tested step count. Even at $N=200$, it surpasses DPS while requiring less runtime. As $N$ increases, runtime scales approximately linearly while memory usage remains nearly constant, and reconstruction quality improves gradually. This smooth quality--efficiency trade-off enables CAT to adapt readily to different inference budgets without increasing memory demand, making it practical for both quality-oriented and resource-constrained settings.

\subsection{Extension}
\label{sec:extension}

\subsubsection{Scientific Inverse Problems}
\label{sec:scientific_inverse_problems}

\paragraph{Evaluation protocol.}
We extend the evaluation beyond natural image restoration to black hole imaging in InverseBench~\cite{zheng2025inversebench}. This task reconstructs a source image from sparse interferometric observations using nonlinear closure-phase and log-closure-amplitude constraints, together with a total-flux constraint. We evaluate DPS+CAT on 100 test images at $64\times64$ resolution using the pretrained black-hole diffusion prior from InverseBench~\cite{zheng2025inversebench} and observation time ratios of $100\%$ and $10\%$. The algorithm hyperparameter settings used in these experiments are provided in Appendix~\ref{app:reproducibility}. We compare against the domain-specific methods SMILI and EHT-Imaging and the diffusion-prior methods DPS, LGD, RED-Diff, PnPDM, DAPS, and DiffPIR, using their reported InverseBench results. Image quality is measured by PSNR and Blur PSNR. We also report the reduced chi-squared statistics $\widetilde\chi^2_{\mathrm{cp}}$ and $\widetilde\chi^2_{\mathrm{logca}}$ for closure phase and log closure amplitude; values near one indicate agreement at the measurement noise level.

\begin{table}[!htbp]
  \centering
  \caption{Black hole imaging on InverseBench at $100\%$ and $10\%$ observation time. Entries are mean (standard deviation). Baseline results are taken from InverseBench~\cite{zheng2025inversebench}; DPS+CAT is evaluated on 100 test images. Bold marks the highest mean PSNR/Blur PSNR and the lowest mean reduced chi-squared within each observation setting.}
  \label{tab:blackhole_results}
  \begingroup
  \scriptsize
  \setlength{\tabcolsep}{3pt}
  \renewcommand{\arraystretch}{1.12}
  \resizebox{\textwidth}{!}{%
  \begin{tabular}{@{}lrrrrrrrr@{}}
    \toprule
    \multirow{2}{*}{\textbf{Method}} & \multicolumn{4}{c}{\textbf{Observation time ratio: 100\%}} & \multicolumn{4}{c}{\textbf{Observation time ratio: 10\%}} \\
    \cmidrule(lr){2-5}\cmidrule(lr){6-9}
    & PSNR $\uparrow$ & Blur PSNR $\uparrow$ & $\widetilde\chi^2_{\mathrm{cp}}\downarrow$ & $\widetilde\chi^2_{\mathrm{logca}}\downarrow$ & PSNR $\uparrow$ & Blur PSNR $\uparrow$ & $\widetilde\chi^2_{\mathrm{cp}}\downarrow$ & $\widetilde\chi^2_{\mathrm{logca}}\downarrow$ \\
    \midrule
    SMILI & 22.67 (3.13) & 27.79 (4.02) & 1.878 (0.952) & 17.612 (10.299) & 20.85 (2.90) & 25.24 (3.86) & 1.209 (0.169) & 21.788 (12.491) \\
    EHT-Imaging & 24.28 (3.63) & 28.57 (4.52) & \textbf{1.251 (0.250)} & 1.259 (0.316) & 22.67 (3.46) & 26.66 (3.93) & \textbf{1.166 (0.156)} & \textbf{1.240 (0.205)} \\
    \midrule
    DPS & 25.86 (3.90) & 32.94 (6.19) & 8.759 (37.784) & 5.456 (24.185) & 24.36 (3.72) & 30.79 (5.75) & 13.052 (43.087) & 6.614 (26.789) \\
    LGD & 21.22 (3.64) & 26.06 (4.98) & 13.239 (17.231) & 13.233 (39.107) & 22.08 (3.75) & 27.48 (5.09) & 10.775 (21.684) & 13.375 (56.397) \\
    RED-Diff & 23.77 (4.13) & 29.13 (6.22) & 1.853 (0.938) & 2.050 (2.361) & 22.53 (3.02) & 27.67 (4.53) & 2.488 (2.925) & 4.916 (13.221) \\
    PnPDM & 26.07 (3.70) & 32.88 (6.02) & 1.311 (0.195) & \textbf{1.199 (0.221)} & 24.57 (3.47) & 30.80 (5.22) & 1.433 (0.417) & 1.336 (0.478) \\
    DAPS & 25.60 (3.64) & 32.78 (5.68) & 1.300 (0.324) & 1.229 (0.532) & 23.99 (3.56) & 30.10 (5.13) & 1.545 (0.705) & 2.253 (9.903) \\
    DiffPIR & 25.01 (4.64) & 31.86 (6.56) & 3.271 (1.623) & 2.970 (1.202) & 23.84 (3.59) & 30.04 (5.03) & 5.374 (3.733) & 5.205 (5.556) \\
    \rowcolor{gray!10}
    DPS+CAT (ours) & \textbf{27.94 (3.95)} & \textbf{35.94 (6.17)} & 2.625 (13.114) & 1.663 (4.784) & \textbf{26.35 (4.04)} & \textbf{33.14 (5.99)} & 1.239 (0.222) & 1.264 (0.221) \\
    \bottomrule
  \end{tabular}%
  }
  \endgroup
\end{table}

\begin{figure}[!htbp]
  \centering
  \begingroup
  \setlength{\tabcolsep}{0.7pt}
  \newcommand{\bhpanel}[1]{%
    \raisebox{-0.5\height}{\includegraphics[width=0.132\textwidth]{#1}}%
  }
  \begin{tabular}{@{}c*{7}{c}@{}}
    & \scriptsize Ground truth & \scriptsize DPS & \scriptsize LGD
    & \scriptsize PnPDM & \scriptsize DAPS & \scriptsize DiffPIR
    & \scriptsize DPS+CAT (ours) \\[-1pt]
    \raisebox{-0.5\height}{\rotatebox{90}{\scriptsize $100\%$}} &
    \bhpanel{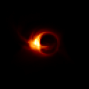} &
    \bhpanel{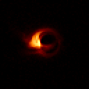} &
    \bhpanel{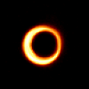} &
    \bhpanel{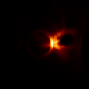} &
    \bhpanel{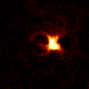} &
    \bhpanel{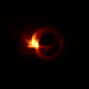} &
    \bhpanel{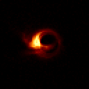} \\
    \noalign{\vskip 5pt}
    \raisebox{-0.5\height}{\rotatebox{90}{\scriptsize $10\%$}} &
    \bhpanel{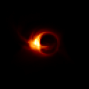} &
    \bhpanel{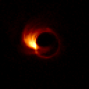} &
    \bhpanel{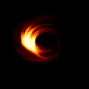} &
    \bhpanel{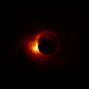} &
    \bhpanel{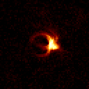} &
    \bhpanel{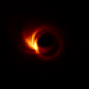} &
    \bhpanel{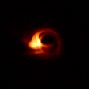}
  \end{tabular}
  \endgroup
  \caption{Qualitative black hole imaging results for the same ground-truth source from InverseBench at $100\%$ (top) and $10\%$ (bottom) observation time.}
  \label{fig:blackhole_qualitative}
\end{figure}

\paragraph{Quantitative comparison.}
Table~\ref{tab:blackhole_results} shows that DPS+CAT consistently achieves the strongest image-space reconstruction quality across both observation-time settings. Compared with the DPS host, CAT improves both PSNR and Blur PSNR, indicating better pixel-wise fidelity and large-scale morphological agreement with the ground truth. This advantage persists under reduced observation time, supporting the applicability of CAT beyond natural-image restoration to a challenging physics-based nonlinear inverse problem. Figure~\ref{fig:blackhole_qualitative} further shows that DPS+CAT recovers a sharper, more continuous emission ring with fewer artifacts, particularly at $10\%$ observation time.

\FloatBarrier

\subsubsection{Classifier-Free Guidance}
\label{sec:cfg_extension}

\paragraph{Evaluation protocol.}
We further apply the curvature-adaptive tubular correction to classifier-free guidance (CFG)~\cite{ho2021classifierfree}. CFG forms a conditional update from the difference between conditional and unconditional model predictions; we pass this guidance direction through CAT before combining it with the unconditional prediction. This construction uses the unconditional score to define the local prior geometry and does not require retraining. We evaluate CFG+CAT on Stable Diffusion 2.1~\cite{rombach2022high} using 1000 captions from the COCO 2017 validation split~\cite{lin2014microsoft}. All methods use the model's official sampler and identical default sampling settings, and same prompts and initial seeds at guidance scale $w\in\{10,20,50,100\}$. We compare with standard CFG and adaptive projected guidance (APG)~\cite{sadat2025eliminating}. Following the same protocol for all methods, we report CLIP score~\cite{radford2021learning}, FID~\cite{heusel2017gans}, saturation, and RMS contrast. CLIP and FID measure text--image alignment and distributional fidelity, respectively, whereas saturation and contrast characterize changes in image appearance under stronger guidance.

\begin{table}[!htbp]
  \centering
  \caption{Classifier-free text-to-image guidance on 1000 COCO 2017 validation captions. Bold marks the best CLIP and FID at each guidance scale.}
  \label{tab:cfg_extension}
  \begingroup
  \small
  \setlength{\tabcolsep}{9pt}
  \renewcommand{\arraystretch}{1.05}
  \begin{tabular}{@{}clrrrr@{}}
    \toprule
    \textbf{Scale $w$} & \textbf{Method} & \textbf{CLIP $\uparrow$} & \textbf{FID $\downarrow$} & \textbf{Sat.} & \textbf{Contrast} \\
    \midrule
    \multirow{3}{*}{10}
      & CFG & \textbf{31.6905} & 66.63 & 0.2901 & 0.2352 \\
      & APG & 31.4236 & 65.26 & 0.2014 & 0.1689 \\
      \rowcolor{gray!10} & CFG+CAT (ours) & 31.2664 & \textbf{64.67} & 0.2228 & 0.1943 \\
    \midrule
    \multirow{3}{*}{20}
      & CFG & \textbf{31.8678} & 69.85 & 0.3538 & 0.2728 \\
      & APG & 31.7346 & 67.04 & 0.2212 & 0.1865 \\
      \rowcolor{gray!10} & CFG+CAT (ours) & 31.2742 & \textbf{65.59} & 0.2263 & 0.1974 \\
    \midrule
    \multirow{3}{*}{50}
      & CFG & 31.6969 & 78.71 & 0.4332 & 0.3199 \\
      & APG & \textbf{31.9653} & 69.71 & 0.3213 & 0.2544 \\
      \rowcolor{gray!10} & CFG+CAT (ours) & 31.3065 & \textbf{65.48} & 0.2285 & 0.1991 \\
    \midrule
    \multirow{3}{*}{100}
      & CFG & 30.6990 & 104.88 & 0.5228 & 0.3489 \\
      & APG & \textbf{32.0276} & 74.59 & 0.4473 & 0.3280 \\
      \rowcolor{gray!10} & CFG+CAT (ours) & 31.3109 & \textbf{66.26} & 0.2295 & 0.1996 \\
    \bottomrule
  \end{tabular}
  \endgroup
\end{table}

\begin{figure}[!htbp]
  \centering
  \begingroup
  \setlength{\tabcolsep}{0.8pt}
  \begin{tabular}{@{}*{6}{c}@{}}
    \scriptsize CFG & \scriptsize APG & \scriptsize CFG+CAT (ours)
    & \scriptsize CFG & \scriptsize APG & \scriptsize CFG+CAT (ours) \\[-1pt]
    \includegraphics[width=0.154\textwidth]{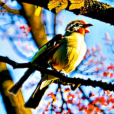} &
    \includegraphics[width=0.154\textwidth]{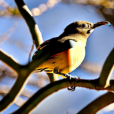} &
    \includegraphics[width=0.154\textwidth]{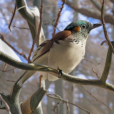} &
    \includegraphics[width=0.154\textwidth]{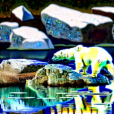} &
    \includegraphics[width=0.154\textwidth]{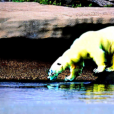} &
    \includegraphics[width=0.154\textwidth]{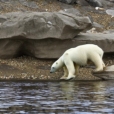}
  \end{tabular}
  \endgroup
  \caption{Qualitative comparison at guidance scale $w=100$. Within each three-image group, all methods use the same prompt and initial seed. From left to right, the groups correspond to ``A bird that is on a tree limb'' and ``A polar bear dives off a rock into a river.''}
  \label{fig:cfg_qualitative}
\end{figure}

\paragraph{Quantitative and qualitative comparison.}
Table~\ref{tab:cfg_extension} shows that CFG+CAT achieves the lowest FID at every evaluated guidance scale while maintaining stable FID, saturation, and contrast as $w$ increases; by comparison, the FID of standard CFG deteriorates substantially, while that of APG increases more moderately. Although CFG+CAT does not attain the highest CLIP score, its alignment remains stable across the sweep. The qualitative results at $w=100$ in Figure~\ref{fig:cfg_qualitative} reinforce these trends: standard CFG produces severe oversaturation and flattened, posterized regions, while APG reduces these artifacts but retains exaggerated color and contrast. In contrast, CFG+CAT yields more coherent layouts, smoother tonal transitions, and more natural colors while preserving prompt-specified content. Together, these results demonstrate that CAT extends beyond inverse-problem guidance and provides a favorable fidelity--alignment trade-off under strong conditional guidance.

\FloatBarrier

\section{Conclusion}
\label{sec:conclusion}

This paper introduced CAT, a training-free correction for stabilizing gradient-guided diffusion. CAT allocates a noise-dependent tubular budget between first-order normal displacement and curvature-induced second-order tangent departure, retaining useful normal guidance while limiting movement beyond locally reliable regions of the prior. The allocation has a unique scalar-dual solution, and Armijo backtracking calibrates each finite step against the guidance objective. Its matrix-free implementation applies in both pixel and latent spaces.

Across seven inverse problems on FFHQ and ImageNet, CAT improved the evaluated DPS, PSLD, and ReSample hosts, particularly in perceptual quality. Ablations and robustness sweeps support the complementary roles of curvature-aware control and finite-step calibration. CAT also improved black hole reconstruction over DPS and achieved the lowest FID among the compared text-to-image guidance methods at every tested scale. Nevertheless, CAT provides local control rather than a guarantee of globally accurate posterior sampling, relies on uncertified local score geometry, and adds curvature-estimation and line-search costs. Adapting its budget to score uncertainty and validating it across broader scientific and conditional-generation settings remain important directions.

\subsection*{Acknowledgments}
Zheng Ma is supported by NSFC Grant No. 12531016 and  Beijing Institute of Applied
Physics and Computational Mathematics funding HX02023-6. Additionally, we also thank Shanghai Institute for Mathematics and Interdisciplinary Sciences (SIMIS) for their financial support. This research was funded by SIMIS under grant number SIMIS-ID-2025-ST. The authors are grateful for the resources and facilities provided by SIMIS, which were essential for the completion of this work.

\clearpage

\bibliographystyle{unsrt}
\bibliography{references}

\appendix
\section{Proofs of the Main Results}
\label{app:proofs}

\subsection{Proof of Theorem~\ref{thm:tubular_expansion}}
\label{app:geometry_proof}

Write $r_t(x):=r(x,\mathcal M_t)$.  At the anchor,
\begin{equation}
  \phi_t(x_t)-c_t=0,\qquad
  D\phi_t(x_t)=s_t^\top,\qquad
  \operatorname{rank}D\phi_t(x_t)=1.
  \label{eq:app_regular_rank}
\end{equation}
The regular-level-set theorem therefore gives
\begin{equation}
  \dim \mathcal M_t=D-1,
  \qquad
  T_{x_t}\mathcal M_t\subseteq\ker(s_t^\top).
  \label{eq:app_tangent_inclusion}
\end{equation}
Indeed, for any $C^1$ curve $\gamma:(-\epsilon,\epsilon)\rightarrow\mathcal M_t$ with $\gamma(0)=x_t$,
\begin{equation}
  0
  =
  \left.
  \frac{d}{d\tau}
  \bigl(\phi_t(\gamma(\tau))-c_t\bigr)
  \right|_{\tau=0}
  =
  s_t^\top\gamma'(0).
  \label{eq:app_tangent_curve}
\end{equation}
Since
\begin{equation}
  \dim T_{x_t}\mathcal M_t
  =
  D-1
  =
  \dim\ker(s_t^\top),
\end{equation}
\eqref{eq:app_tangent_inclusion} is an equality:
\begin{equation}
  T_{x_t}\mathcal M_t
  =
  \ker(s_t^\top)
  =
  \{v\in\mathbb R^D:s_t^\top v=0\}.
  \label{eq:app_tangent_space}
\end{equation}
Moreover,
\begin{equation}
  \|\nu_t\|_2=1,\qquad
  P_{N,t}^2=P_{N,t},\qquad
  P_{T,t}^2=P_{T,t},\qquad
  P_{N,t}P_{T,t}=0,\qquad
  P_{N,t}+P_{T,t}=I.
  \label{eq:app_projector_identities}
\end{equation}

Extend the score and score-norm notation from \eqref{eq:local_geometry} to nearby
points by
\begin{equation}
  s_t(x):=\nabla\phi_t(x),\qquad
  g_t(x):=\|s_t(x)\|_2,\qquad
  \nu_t(x)=\frac{s_t(x)}{g_t(x)},
\end{equation}
so that $s_t(x_t)=s_t$ and $g_t(x_t)=g_t$.  For every
$v\in\mathbb R^D$,
\begin{equation}
  Ds_t(x_t)[v]=H_tv.
  \label{eq:app_score_derivative}
\end{equation}
Differentiating $g_t(x)^2=s_t(x)^\top s_t(x)$ gives
\begin{equation}
\begin{aligned}
  2g_t(x_t)Dg_t(x_t)[v]
  &=
  2s_t(x_t)^\top Ds_t(x_t)[v],\\
  Dg_t(x_t)[v]
  &=
  \frac{s_t^\top H_tv}{g_t}
  =
  \nu_t^\top H_tv.
\end{aligned}
\label{eq:app_norm_derivative}
\end{equation}
Hence
\begin{equation}
\begin{aligned}
  D\nu_t(x_t)[v]
  &=
  \frac{Ds_t(x_t)[v]}{g_t}
  -
  \frac{s_t}{g_t^2}Dg_t(x_t)[v]\\
  &=
  \frac{1}{g_t}
  \left(
    H_tv-\nu_t\nu_t^\top H_tv
  \right)\\
  &=
  \frac{1}{g_t}P_{T,t}H_tv,
\end{aligned}
\label{eq:app_normal_derivative}
\end{equation}
Projecting the argument and the result onto the tangent space yields
\begin{equation}
  P_{T,t}D\nu_t(x_t)P_{T,t}
  =
  \frac{1}{g_t}P_{T,t}H_tP_{T,t}
  =
  B_t.
  \label{eq:app_curvature_operator}
\end{equation}
Thus, for $u=P_{T,t}u$ and $\|u\|_2=1$,
\begin{equation}
  \kappa_t(u)
  =
  u^\top D\nu_t(x_t)[u]
  =
  u^\top B_tu
  =
  \frac{u^\top H_tu}{g_t},
  \label{eq:app_directional_curvature}
\end{equation}
which proves \eqref{eq:curvature_matrix}.

On $\mathcal M_t$ and throughout its tubular neighborhood,
\begin{equation}
  r_t|_{\mathcal M_t}=0,\qquad
  \nabla r_t(z)=\nu_t(z)\quad(z\in\mathcal M_t),\qquad
  \|\nabla r_t(x)\|_2^2=1.
  \label{eq:app_distance_identities}
\end{equation}
For every $w\in\mathbb R^D$,
\begin{equation}
\begin{aligned}
  0
  &=
  \frac{1}{2}
  D\!\left(
    \|\nabla r_t(x)\|_2^2
  \right)[w]\\
  &=
  w^\top\nabla^2r_t(x)\nabla r_t(x).
\end{aligned}
\end{equation}
Since this holds for all $w$,
\begin{equation}
  \nabla^2r_t(x)\nabla r_t(x)=0,
  \qquad
  \nabla^2r_t(x_t)\nu_t=0,
  \qquad
  \nu_t^\top\nabla^2r_t(x_t)=0.
  \label{eq:app_eikonal_hessian}
\end{equation}
For $v\in T_{x_t}\mathcal M_t$, take $\gamma(0)=x_t$ and $\gamma'(0)=v$ in \eqref{eq:app_tangent_curve}.  Differentiating $\nabla r_t(\gamma(\tau))=\nu_t(\gamma(\tau))$ at $\tau=0$ gives
\begin{equation}
  \nabla^2r_t(x_t)v
  =
  D\nu_t(x_t)[v].
  \label{eq:app_distance_tangent_hessian}
\end{equation}
Equations \eqref{eq:app_curvature_operator}, \eqref{eq:app_eikonal_hessian}, and \eqref{eq:app_projector_identities} imply
\begin{equation}
\begin{aligned}
  \nabla^2r_t(x_t)
  &=
  (P_{N,t}+P_{T,t})
  \nabla^2r_t(x_t)
  (P_{N,t}+P_{T,t})\\
  &=
  P_{T,t}\nabla^2r_t(x_t)P_{T,t}\\
  &=
  \frac{1}{g_t}P_{T,t}H_tP_{T,t}
  =
  B_t.
\end{aligned}
\label{eq:app_distance_hessian}
\end{equation}

For a displacement $d$ whose segment $\{x_t+\tau d:0\leq\tau\leq1\}$ lies in $\mathcal U_t$, define
\begin{equation}
  \psi_d(\tau):=r_t(x_t+\tau d).
\end{equation}
Then
\begin{equation}
\begin{aligned}
  \psi_d(0)&=0,\\
  \psi_d'(0)
  &=
  \nabla r_t(x_t)^\top d
  =
  \nu_t^\top(d_N+d_T)
  =
  \nu_t^\top d_N,\\
  \psi_d''(0)
  &=
  d^\top\nabla^2r_t(x_t)d
  =
  d^\top B_td
  =
  d_T^\top B_td_T,\\
  \psi_d'''(\tau)
  &=
  D^3r_t(x_t+\tau d)[d,d,d].
\end{aligned}
\label{eq:app_taylor_derivatives}
\end{equation}
Taylor's formula with integral remainder gives
\begin{equation}
\begin{aligned}
  r_t(x_t+d)
  &=
  \psi_d(1)\\
  &=
  \psi_d(0)+\psi_d'(0)+\frac{1}{2}\psi_d''(0)+R_3(d)\\
  &=
  \nu_t^\top d_N
  +\frac{1}{2}d_T^\top B_td_T
  +R_3(d),
\end{aligned}
\label{eq:app_taylor_expansion}
\end{equation}
where
\begin{equation}
  R_3(d)
  =
  \frac{1}{2}
  \int_0^1
  (1-\tau)^2
  D^3r_t(x_t+\tau d)[d,d,d]\,d\tau.
  \label{eq:app_integral_remainder}
\end{equation}
Therefore,
\begin{equation}
\begin{aligned}
  |R_3(d)|
  &\leq
  \frac{1}{2}
  \int_0^1
  (1-\tau)^2
  \|D^3r_t(x_t+\tau d)\|_{\mathrm{op}}
  \|d\|_2^3\,d\tau\\
  &\leq
  \frac{M_3}{2}\|d\|_2^3
  \int_0^1(1-\tau)^2\,d\tau\\
  &=
  \frac{M_3}{6}\|d\|_2^3.
\end{aligned}
\label{eq:app_remainder_bound}
\end{equation}
Equations \eqref{eq:app_taylor_expansion} and \eqref{eq:app_remainder_bound} prove \eqref{eq:tubular_expansion}.

\subsection{Proof of Theorem~\ref{thm:kkt_solution}}
\label{app:kkt_proof}

For notational brevity, set
\begin{equation}
\begin{aligned}
  m_t(r_N,r_T)
  &:=
  -a_tr_N-b_tr_T
  +\frac{r_N^2+r_T^2}{2\alpha_t},\\
  c_t(r_N,r_T)
  &:=
  r_N+\frac{1}{2}K_tr_T^2-R_t,\\
  \mathcal C_t
  &:=
  \{(r_N,r_T)\in\mathbb R_+^2:c_t(r_N,r_T)\leq0\}.
\end{aligned}
\label{eq:app_primal_definitions}
\end{equation}
Their Hessians are
\begin{equation}
  \nabla^2m_t
  =
  \frac{1}{\alpha_t}I_2
  \succ0,
  \qquad
  \nabla^2c_t
  =
  \begin{bmatrix}
    0&0\\
    0&K_t
  \end{bmatrix}
  \succeq0.
  \label{eq:app_primal_hessians}
\end{equation}
Thus $m_t$ is strictly convex and $\mathcal C_t$ is convex.  Moreover,
\begin{equation}
\begin{aligned}
  m_t(r_N,r_T)
  &=
  \frac{1}{2\alpha_t}
  \left\|
    \begin{bmatrix}r_N\\r_T\end{bmatrix}
    -
    \alpha_t
    \begin{bmatrix}a_t\\b_t\end{bmatrix}
  \right\|_2^2
  -
  \frac{\alpha_t}{2}(a_t^2+b_t^2),
\end{aligned}
\label{eq:app_primal_square}
\end{equation}
so
\begin{equation}
  \lim_{\|(r_N,r_T)\|_2\rightarrow\infty}
  m_t(r_N,r_T)
  =
  +\infty.
  \label{eq:app_coercivity}
\end{equation}
The set $\mathcal C_t$ is nonempty because $(0,0)\in\mathcal C_t$, and it is closed because $c_t$ is continuous.  For $K_t>0$, choose
\begin{equation}
  0<\epsilon
  <
  \min\left\{
    \frac{R_t}{2},
    \sqrt{\frac{R_t}{K_t}}
  \right\};
\end{equation}
then
\begin{equation}
  \epsilon+\frac{1}{2}K_t\epsilon^2
  <
  \frac{R_t}{2}+\frac{R_t}{2}
  =
  R_t.
\end{equation}
For $K_t=0$, any $0<\epsilon<R_t$ gives the same strict inequality. Consequently,
\begin{equation}
  c_t(\epsilon,\epsilon)<0,\qquad
  -\epsilon<0,\qquad
  -\epsilon<0,
  \label{eq:app_slater_point}
\end{equation}
For multipliers $\lambda_t,\mu_N,\mu_T\geq0$, define
\begin{equation}
\begin{aligned}
  \mathcal L
  ={}&
  -a_tr_N-b_tr_T
  +\frac{r_N^2+r_T^2}{2\alpha_t}\\
  &+
  \lambda_t
  \left(
    r_N+\frac{1}{2}K_tr_T^2-R_t
  \right)
  -\mu_Nr_N-\mu_Tr_T .
\end{aligned}
\label{eq:app_lagrangian}
\end{equation}
Slater's condition holds, so equations \eqref{eq:app_primal_hessians}--\eqref{eq:app_slater_point} imply
\begin{equation}
\begin{aligned}
  \operatorname*{arg\,min}_{(r_N,r_T)\in\mathcal C_t}
  m_t(r_N,r_T)
  &=
  \{(r_{N,t}^*,r_{T,t}^*)\},\\
  \min_{(r_N,r_T)\in\mathcal C_t}m_t(r_N,r_T)
  &=
  \max_{\lambda_t,\mu_N,\mu_T\geq0}
  \inf_{(r_N,r_T)\in\mathbb R^2}
  \mathcal L(r_N,r_T,\lambda_t,\mu_N,\mu_T),\\
  (r_N,r_T)\text{ is optimal}
  &\Longleftrightarrow
  (r_N,r_T,\lambda_t,\mu_N,\mu_T)
  \text{ satisfies the KKT system}.
\end{aligned}
  \label{eq:app_convex_conclusions}
\end{equation}

The complete KKT system is
\begin{equation}
\left\{
\begin{aligned}
  -a_t+\frac{r_N}{\alpha_t}+\lambda_t-\mu_N&=0,\\
  -b_t+\frac{r_T}{\alpha_t}+\lambda_tK_tr_T-\mu_T&=0,\\
  r_N+\frac{1}{2}K_tr_T^2-R_t&\leq0,\\
  r_N,\ r_T,\ \lambda_t,\ \mu_N,\ \mu_T&\geq0,\\
  \lambda_t
  \left(
    r_N+\frac{1}{2}K_tr_T^2-R_t
  \right)&=0,\\
  \mu_Nr_N=\mu_Tr_T&=0.
\end{aligned}
\right.
\label{eq:app_complete_kkt}
\end{equation}

The normal stationarity and complementarity conditions give
\begin{equation}
  \begin{cases}
    r_N>0
    \ \Longrightarrow\
    \mu_N=0,\ 
    r_N=\alpha_t(a_t-\lambda_t),\
    \lambda_t<a_t,\\[2mm]
    r_N=0
    \ \Longrightarrow\
    \mu_N=\lambda_t-a_t\geq0,\
    \lambda_t\geq a_t.
  \end{cases}
\end{equation}
Hence
\begin{equation}
  r_N(\lambda_t)
  =
  \alpha_t(a_t-\lambda_t)_+.
  \label{eq:app_normal_solution}
\end{equation}
If $b_t>0$ and $r_T=0$, tangent stationarity would require
\begin{equation}
  -b_t-\mu_T=0
  \quad\Longrightarrow\quad
  \mu_T=-b_t<0,
\end{equation}
contradicting dual feasibility.  Therefore $r_T>0$, $\mu_T=0$, and
\begin{equation}
  \left(
    \frac{1}{\alpha_t}+\lambda_tK_t
  \right)r_T
  =
  b_t
  \quad\Longrightarrow\quad
  r_T(\lambda_t)
  =
  \frac{\alpha_tb_t}
       {1+\alpha_t\lambda_tK_t}.
  \label{eq:app_tangent_solution}
\end{equation}
If $b_t=0$, tangent stationarity and complementarity give
\begin{equation}
  r_T
  \left[
    \left(
      \frac{1}{\alpha_t}+\lambda_tK_t
    \right)r_T-\mu_T
  \right]
  =
  \left(
    \frac{1}{\alpha_t}+\lambda_tK_t
  \right)r_T^2
  =
  0
  \quad\Longrightarrow\quad
  r_T=0,
\end{equation}
which is also \eqref{eq:app_tangent_solution}.  Equations \eqref{eq:app_normal_solution} and \eqref{eq:app_tangent_solution} therefore prove the claimed closed-form expressions for $r_{N,t}(\lambda_t)$ and $r_{T,t}(\lambda_t)$.

At $\lambda_t=0$,
\begin{equation}
  (r_N(0),r_T(0))
  =
  (\alpha_ta_t,\alpha_tb_t),
\end{equation}
and its constraint value is
\begin{equation}
  F_t(0)-R_t
  =
  \alpha_ta_t
  +\frac{1}{2}K_t(\alpha_tb_t)^2
  -R_t.
  \label{eq:app_unconstrained_constraint}
\end{equation}
If $\alpha_ta_t+\frac{1}{2}K_t(\alpha_tb_t)^2\leq R_t$, then $\lambda_t=0$ together with \eqref{eq:app_normal_solution}--\eqref{eq:app_tangent_solution} satisfies \eqref{eq:app_complete_kkt}, and hence gives the unique optimum.

If this condition fails, then $\lambda_t\neq0$ by \eqref{eq:app_unconstrained_constraint}. Thus $\lambda_t>0$, and complementarity enforces
\begin{equation}
  r_N(\lambda_t)
  +\frac{1}{2}K_tr_T(\lambda_t)^2
  =
  R_t,
\end{equation}
which is exactly \eqref{eq:dual_equation}.

For $\lambda\geq0$, define
\begin{equation}
  F_t(\lambda)
  =
  \alpha_t(a_t-\lambda)_+
  +
  \frac{K_t\alpha_t^2b_t^2}
       {2(1+\alpha_t\lambda K_t)^2}.
\end{equation}
It satisfies
\begin{equation}
  F_t(0)>R_t>0,
  \qquad
  \lim_{\lambda\rightarrow\infty}F_t(\lambda)=0<R_t.
  \label{eq:app_root_endpoints}
\end{equation}
On $\lambda<a_t$,
\begin{equation}
  F_t'(\lambda)
  =
  -\alpha_t
  -
  \frac{K_t^2\alpha_t^3b_t^2}
       {(1+\alpha_t\lambda K_t)^3}
  <
  0.
  \label{eq:app_root_derivative_below}
\end{equation}
On $\lambda>a_t$,
\begin{equation}
  F_t'(\lambda)
  =
  -
  \frac{K_t^2\alpha_t^3b_t^2}
       {(1+\alpha_t\lambda K_t)^3}
  \leq0.
  \label{eq:app_root_derivative_above}
\end{equation}
Continuity and \eqref{eq:app_root_endpoints} give existence of a root.  For $0\leq\lambda_1<\lambda_2$ with $F_t(\lambda_1)>0$,
\begin{equation}
  \begin{cases}
    \lambda_1<a_t
    &\Longrightarrow\
    F_t(\lambda_2)<F_t(\lambda_1),\\
    \lambda_1\geq a_t
    &\Longrightarrow\
    K_tb_t>0
    \Longrightarrow\
    F_t(\lambda_2)<F_t(\lambda_1).
  \end{cases}
  \label{eq:app_root_strictness}
\end{equation}
Therefore two distinct roots at the positive level $R_t$ cannot exist, so the nonnegative solution of \eqref{eq:dual_equation} is unique.

\subsection{Proof of Theorem~\ref{thm:descent_armijo}}
\label{app:descent_proof}

For any $d$ whose update segment is in the $L_g$-smooth neighborhood,
\begin{equation}
\begin{aligned}
  L_t(x_t+d)-L_t(x_t)
  &=
  \int_0^1
  \nabla L_t(x_t+\tau d)^\top d\,d\tau\\
  &=
  q_t^\top d
  +
  \int_0^1
  \left[
    \nabla L_t(x_t+\tau d)-\nabla L_t(x_t)
  \right]^\top d\,d\tau\\
  &\leq
  q_t^\top d
  +
  \int_0^1
  \left\|
    \nabla L_t(x_t+\tau d)-\nabla L_t(x_t)
  \right\|_2
  \|d\|_2\,d\tau\\
  &\leq
  q_t^\top d
  +
  \int_0^1
  L_g\tau\|d\|_2^2\,d\tau\\
  &=
  q_t^\top d+\frac{L_g}{2}\|d\|_2^2.
\end{aligned}
\label{eq:descent_lemma_appendix}
\end{equation}

Let
\begin{equation}
  d_t^*
  =
  -r_{N,t}^*e_{N,t}-r_{T,t}^*e_{T,t}.
\end{equation}
Orthogonality of the normal and tangent components gives
\begin{equation}
\begin{aligned}
  q_t^\top d_t^*
  &=
  (q_{N,t}+q_{T,t})^\top
  (-r_{N,t}^*e_{N,t}-r_{T,t}^*e_{T,t})\\
  &=
  -a_tr_{N,t}^*-b_tr_{T,t}^*,\\
  \|d_t^*\|_2^2
  &=
  (r_{N,t}^*)^2+(r_{T,t}^*)^2.
\end{aligned}
\label{eq:app_direction_identities}
\end{equation}
Since $(0,0)$ is feasible,
\begin{equation}
  m_t(r_{N,t}^*,r_{T,t}^*)
  \leq
  m_t(0,0)
  =
  0.
  \label{eq:proximal_nonpositive}
\end{equation}
If $a_t>0$, choose
\begin{equation}
  0<\epsilon_N<\min\{R_t,2\alpha_ta_t\}.
\end{equation}
Then $(\epsilon_N,0)$ is feasible and
\begin{equation}
  m_t(\epsilon_N,0)
  =
  -a_t\epsilon_N
  +\frac{\epsilon_N^2}{2\alpha_t}
  =
  -\epsilon_N
  \left(
    a_t-\frac{\epsilon_N}{2\alpha_t}
  \right)
  <
  0.
  \label{eq:app_strict_normal_test}
\end{equation}
If $a_t=0$ and $b_t>0$, select
\begin{equation}
  0<\epsilon_T
  <
  \begin{cases}
    \min\!\left\{
      2\alpha_tb_t,\sqrt{2R_t/K_t}
    \right\},
    &K_t>0,\\
    2\alpha_tb_t,
    &K_t=0.
  \end{cases}
\end{equation}
Then $(0,\epsilon_T)$ is feasible and
\begin{equation}
  m_t(0,\epsilon_T)
  =
  -b_t\epsilon_T
  +\frac{\epsilon_T^2}{2\alpha_t}
  =
  -\epsilon_T
  \left(
    b_t-\frac{\epsilon_T}{2\alpha_t}
  \right)
  <
  0.
  \label{eq:app_strict_tangent_test}
\end{equation}
Thus
\begin{equation}
  q_t\neq0
  \quad\Longrightarrow\quad
  a_t>0\ \text{or}\ b_t>0
  \quad\Longrightarrow\quad
  m_t(r_{N,t}^*,r_{T,t}^*)<0.
  \label{eq:app_strict_optimal_value}
\end{equation}
Using \eqref{eq:app_direction_identities},
\begin{equation}
  m_t(r_{N,t}^*,r_{T,t}^*)
  =
  q_t^\top d_t^*
  +
  \frac{1}{2\alpha_t}\|d_t^*\|_2^2.
  \label{eq:app_proximal_direction_form}
\end{equation}
For $0<\alpha_t\leq1/L_g$,
\begin{equation}
\begin{aligned}
  L_t(x_t+d_t^*)-L_t(x_t)
  &\leq
  q_t^\top d_t^*
  +\frac{L_g}{2}\|d_t^*\|_2^2\\
  &\leq
  q_t^\top d_t^*
  +\frac{1}{2\alpha_t}\|d_t^*\|_2^2\\
  &=
  m_t(r_{N,t}^*,r_{T,t}^*)\\
  &\leq0.
\end{aligned}
\label{eq:app_small_step_chain}
\end{equation}
Equations \eqref{eq:app_strict_optimal_value} and \eqref{eq:app_small_step_chain} make the final inequality strict when $q_t\neq0$, proving the stated full-step decrease.

For arbitrary $\alpha_t>0$, \eqref{eq:app_strict_optimal_value} and \eqref{eq:app_proximal_direction_form} imply
\begin{equation}
  q_t^\top d_t^*
  <
  -\frac{1}{2\alpha_t}\|d_t^*\|_2^2
  <
  0
  \qquad(q_t\neq0).
  \label{eq:app_strict_descent_direction}
\end{equation}
For $\gamma\in(0,1]$, apply \eqref{eq:descent_lemma_appendix} to $\gamma d_t^*$:
\begin{equation}
  L_t(x_t+\gamma d_t^*)-L_t(x_t)
  \leq
  \gamma q_t^\top d_t^*
  +\frac{L_g}{2}\gamma^2\|d_t^*\|_2^2.
  \label{eq:app_scaled_descent}
\end{equation}
The Armijo inequality follows whenever
\begin{equation}
\begin{aligned}
  \gamma q_t^\top d_t^*
  +\frac{L_g}{2}\gamma^2\|d_t^*\|_2^2
  &\leq
  c\gamma q_t^\top d_t^*\\
  \Longleftrightarrow\qquad
  \gamma
  &\leq
  \overline\gamma_t
  :=
  \frac{
    2(1-c)(-q_t^\top d_t^*)
  }{
    L_g\|d_t^*\|_2^2
  }.
\end{aligned}
\label{eq:app_armijo_threshold}
\end{equation}
Define
\begin{equation}
  \gamma_{\mathrm{req}}
  :=
  \min\{1,\overline\gamma_t\},
  \qquad
  j_\star
  :=
  \max\left\{
    0,
    \left\lceil
      \frac{\log\gamma_{\mathrm{req}}}{\log\beta}
    \right\rceil
  \right\}.
  \label{eq:app_backtracking_bound}
\end{equation}
Because $0<\beta<1$,
\begin{equation}
  0<\beta^{j_\star}
  \leq
  \gamma_{\mathrm{req}}
  \leq
  \overline\gamma_t.
\end{equation}
Hence Armijo backtracking accepts no later than iteration $j_\star$, and
\begin{equation}
\begin{aligned}
  L_t(x_t+\beta^{j_\star}d_t^*)
  &\leq
  L_t(x_t)
  +c\beta^{j_\star}q_t^\top d_t^*\\
  &<
  L_t(x_t),
\end{aligned}
\end{equation}
which proves \eqref{eq:armijo_condition}.

Finally, for any accepted $\gamma_t\in(0,1]$,
\begin{equation}
\begin{aligned}
  \gamma_tr_{N,t}^*
  +\frac{1}{2}K_t\gamma_t^2(r_{T,t}^*)^2
  &\leq
  \gamma_t
  \left[
    r_{N,t}^*
    +\frac{1}{2}K_t(r_{T,t}^*)^2
  \right]\\
  &\leq
  \gamma_tR_t
  \leq
  R_t,
\end{aligned}
\end{equation}
which proves the stated second-order feasibility.

\section{Reproducibility and Hyperparameter Settings}
\label{app:reproducibility}

\paragraph{Inverse-problem operators.}
The natural-image benchmark uses random-mask inpainting with a missing-pixel probability of $0.7$, box inpainting with a $128\times128$ mask, $4\times$ super-resolution, motion and Gaussian deblurring with $61\times61$ kernels and intensities $0.5$ and $3.0$, respectively, high-dynamic-range reconstruction, and nonlinear deblurring. Gaussian measurement noise with standard deviation $\sigma_y=0.05$ is added in every task except the noise-robustness experiment, which evaluates multiple noise levels.

\paragraph{Configuration reporting protocol.}
Tables~\ref{tab:dps_cat_hyperparameters}--\ref{tab:resample_cat_hyperparameters} summarize the hyperparameters used in the natural-image experiments reported in Table~\ref{tab:main_results}; Table~\ref{tab:dps_cat_hyperparameters} additionally reports the DPS+CAT configuration used for the scientific inverse problem in Table~\ref{tab:blackhole_results} and the CFG+CAT configuration used in Table~\ref{tab:cfg_extension}.  For the inverse-problem integrations, $\rho$ controls the noise-dependent tubular budget $R_t=\rho\sigma_t$, $p$ denotes the interval between Armijo evaluations, and $N$ is the number of reverse sampling steps.  We use the Armijo constants $c=10^{-4}$ and $\beta=0.5$ throughout, with at most three backtracking trials whenever this limit is configurable.  Unless otherwise reported in these tables, the hyperparameters of DPS+CAT, PSLD+CAT, and Resample+CAT are kept identical to those of DPS~\cite{chung2023diffusion}, PSLD~\cite{rout2023solving}, and ReSample~\cite{song2024solving}, respectively.

\paragraph{Pretrained model provenance.}
For the pixel-space experiments, the FFHQ prior is the pretrained diffusion model released with DPS by Chung~et~al.~\cite{chung2023diffusion}, and the ImageNet prior is the unconditional $256\times256$ diffusion model of Dhariwal and Nichol~\cite{dhariwal2021diffusion}.  For the latent-space experiments, the FFHQ prior is the public LDM-VQ-4 checkpoint and the ImageNet prior is the public class-conditional LDM-VQ-8 checkpoint released by Rombach~et~al.~\cite{rombach2022high}.  The scientific extension uses the public $64\times64$ black-hole diffusion prior released with InverseBench~\cite{zheng2025inversebench}. For CFG+CAT, we use the public Stable Diffusion 2.1 checkpoint \texttt{sd2-community/stable-diffusion-2-1}\footnote{\url{https://huggingface.co/sd2-community/stable-diffusion-2-1}} at its native $768\times768$ resolution. All diffusion priors are used without retraining or task-specific fine-tuning.

\begin{table}[!htbp]
  \centering
  \caption{DPS+CAT and CFG+CAT hyperparameters used in the natural-image, scientific inverse-problem, and text-to-image experiments. The host guidance step corresponds to \texttt{scale} for natural images, \texttt{step\_size} for InverseBench, and the CFG scale $w$ for COCO. The symbols $\rho$, $p$, and $N$ denote the tubular-radius coefficient, Armijo period, and denoising-step count, respectively. All CFG+CAT runs use a score-norm threshold of $10^{-8}$ and 30 bisection iterations with tolerance $10^{-4}$.}
  \label{tab:dps_cat_hyperparameters}
  \begingroup
  \small
  \setlength{\tabcolsep}{8pt}
  \renewcommand{\arraystretch}{1.08}
  \begin{tabular}{@{}llrrrr@{}}
    \toprule
    \textbf{Dataset} & \textbf{Inverse problem} & \textbf{Host step} & $\boldsymbol{\rho}$ & $\boldsymbol{p}$ & $\boldsymbol{N}$ \\
    \midrule
    \multirow{7}{*}{FFHQ}
      & Random inpainting    & $4.0$  & $0.1$  & $1$ & $1000$ \\
      & Box inpainting       & $2.0$  & $0.1$  & $1$ & $1000$ \\
      & Super-resolution     & $2.0$  & $0.1$  & $1$ & $1000$ \\
      & Motion deblurring    & $5.0$  & $0.1$  & $1$ & $1000$ \\
      & Gaussian deblurring  & $5.0$  & $0.1$  & $1$ & $1000$ \\
      & HDR reconstruction   & $10.0$ & $0.1$  & $1$ & $1000$ \\
      & Nonlinear deblurring & $5.0$  & $0.1$  & $1$ & $1000$ \\
    \midrule
    \multirow{7}{*}{ImageNet}
      & Random inpainting    & $2.0$  & $1.0$  & $1$ & $1000$ \\
      & Box inpainting       & $4.0$  & $1.0$  & $1$ & $1000$ \\
      & Super-resolution     & $1.5$  & $1.0$  & $1$ & $1000$ \\
      & Motion deblurring    & $6.0$  & $1.0$  & $1$ & $1000$ \\
      & Gaussian deblurring  & $5.0$  & $1.0$  & $1$ & $1000$ \\
      & HDR reconstruction   & $5.0$  & $0.01$ & $1$ & $1000$ \\
      & Nonlinear deblurring & $10.0$ & $1.0$  & $1$ & $1000$ \\
    \midrule
    \multirow{2}{*}{InverseBench}
      & Black-hole imaging ($100\%$ obs.) & $95$ & $0.0005$ & $1$ & $1000$ \\
      & Black-hole imaging ($10\%$ obs.)  & $95$ & $0.0005$ & $1$ & $1000$ \\
    \midrule
    \multirow{1}{*}{COCO 2017}
      & \multirow{1}{*}{Text-to-image} & $10/20/50/100$  & $1.0$ & 1 & $50$ \\
    \bottomrule
  \end{tabular}
  \endgroup
\end{table}
\FloatBarrier

\begin{table}[!htbp]
  \centering
  \caption{PSLD+CAT hyperparameters used in the main experiments.  The original PSLD hyperparameters are the latent data-consistency weight $\gamma$ and guidance weight $\omega$; the remaining notation follows Table~\ref{tab:dps_cat_hyperparameters}.  PSLD is evaluated on the five linear inverse problems supported by its host implementation.}
  \label{tab:psld_cat_hyperparameters}
  \begingroup
  \small
  \setlength{\tabcolsep}{7pt}
  \renewcommand{\arraystretch}{1.10}
  \begin{tabular}{@{}llrrrrr@{}}
    \toprule
    \textbf{Dataset} & \textbf{Inverse problem} & $\boldsymbol{\gamma}$ & $\boldsymbol{\omega}$ & $\boldsymbol{\rho}$ & $\boldsymbol{p}$ & $\boldsymbol{N}$ \\
    \midrule
    \multirow{5}{*}{FFHQ}
      & Random inpainting   & $0.01$ & $2.0$  & $1.0$ & $1$ & $1000$ \\
      & Box inpainting      & $0.01$ & $0.5$  & $1.0$ & $1$ & $1000$ \\
      & Super-resolution    & $0.01$ & $5.0$  & $1.0$ & $1$ & $1000$ \\
      & Motion deblurring   & $0.01$ & $5.0$  & $1.0$ & $1$ & $1000$ \\
      & Gaussian deblurring & $0.10$ & $10.0$ & $1.0$ & $1$ & $1000$ \\
    \midrule
    \multirow{5}{*}{ImageNet}
      & Random inpainting   & $0.10$ & $4.0$ & $0.1$ & $1$ & $1000$ \\
      & Box inpainting      & $0.01$ & $1.0$ & $0.1$ & $1$ & $1000$ \\
      & Super-resolution    & $0.10$ & $2.0$ & $0.1$ & $1$ & $1000$ \\
      & Motion deblurring   & $0.10$ & $1.0$ & $0.1$ & $1$ & $1000$ \\
      & Gaussian deblurring & $0.10$ & $2.0$ & $0.1$ & $1$ & $1000$ \\
    \bottomrule
  \end{tabular}
  \endgroup
\end{table}
\FloatBarrier

\begin{table}[!htbp]
  \centering
  \caption{Resample+CAT hyperparameters used in the main experiments.  $s_{\mathrm{DPS}}$ is the scale of the soft latent DPS step, $P_{\mathrm{tt}}$ is the time-travel period, and the remaining notation follows Table~\ref{tab:dps_cat_hyperparameters}.}
  \label{tab:resample_cat_hyperparameters}
  \begingroup
  \small
  \setlength{\tabcolsep}{5.5pt}
  \renewcommand{\arraystretch}{1.08}
  \begin{tabular}{@{}llrrrrr@{}}
    \toprule
    \textbf{Dataset} & \textbf{Inverse problem} & $\boldsymbol{s_{\mathrm{DPS}}}$ & $\boldsymbol{P_{\mathrm{tt}}}$ & $\boldsymbol{\rho}$ & $\boldsymbol{p}$ & $\boldsymbol{N}$ \\
    \midrule
    \multirow{7}{*}{FFHQ}
      & Random inpainting    & $15.0$  & $10$  & $0.1$ & $5$ & $500$ \\
      & Box inpainting       & $5.0$   & $500$ & $0.1$ & $5$ & $500$ \\
      & Super-resolution     & $150.0$ & $500$ & $0.1$ & $5$ & $500$ \\
      & Motion deblurring    & $5.0$   & $10$  & $0.1$ & $5$ & $500$ \\
      & Gaussian deblurring  & $10.0$  & $10$  & $0.1$ & $5$ & $500$ \\
      & HDR reconstruction   & $30.0$  & $500$ & $0.1$ & $1$ & $500$ \\
      & Nonlinear deblurring & $1.0$   & $10$  & $0.1$ & $1$ & $500$ \\
    \midrule
    \multirow{5}{*}{ImageNet}
      & Random inpainting    & $2.0$  & $10$ & $0.1$ & $5$ & $500$ \\
      & Box inpainting       & $2.0$  & $10$ & $0.1$ & $5$ & $500$ \\
      & Super-resolution     & $15.0$ & $10$ & $0.1$ & $5$ & $500$ \\
      & Motion deblurring    & $2.5$  & $10$ & $0.1$ & $5$ & $500$ \\
      & Gaussian deblurring  & $2.0$  & $10$ & $0.1$ & $5$ & $500$ \\
    \bottomrule
  \end{tabular}
  \endgroup
\end{table}
\FloatBarrier

\paragraph{Settings of the comparison methods.}
For DPS, PSLD, ReSample, DDRM, DDNM, DAPS, DCDP, DiffRGD, DPS+Diffstate, PSLD+DiffState, and Resample+Diffstate, we use the hyperparameter settings reported in their respective original papers whenever available and otherwise tune the unspecified hyperparameters by grid search.

\end{document}